%% file: acl_latex.tex
\documentclass[11pt]{article}

\usepackage[final]{acl}

\usepackage{times}
\usepackage{latexsym}

\usepackage[T1]{fontenc}

\usepackage[utf8]{inputenc}

\usepackage{microtype}

\usepackage{inconsolata}

\usepackage{graphicx}

\title{AdaSearch: Balancing Parametric Knowledge and Search in \\Large Language Models via Reinforcement Learning}

\author{Tzu-Han Lin$^\star$\quad Wei-Lin Chen$^\dag$\quad Chen-An Li$^\star$\\
\textbf{Hung-yi Lee}$^\star$\quad \textbf{Yun-Nung Chen}$^\star$\quad \textbf{Yu Meng}$^\dag$ \\
$^\star$National Taiwan University, Taipei, Taiwan \\ $^\dag$University of Virginia, Charlottesville, VA, USA \\
\texttt{\{r12944034,f13942069,hungyilee\}@ntu.edu.tw} \quad 
\texttt{y.v.chen@ieee.org}\\
\texttt{\{wlchen,yumeng5\}@virginia.edu}
}

\usepackage{subcaption}
\usepackage{hyperref}
\usepackage{booktabs}
\usepackage{amsmath}
\usepackage{amssymb}
\usepackage{mathtools}
\usepackage{amsthm}
\usepackage{xspace}
\usepackage{multirow}
\usepackage[table]{xcolor}
\usepackage[dvipsnames]{xcolor}
\usepackage{dsfont}
\usepackage{placeins}
\usepackage{enumitem}
\usepackage{xurl}
\usepackage{arydshln}
\usepackage{algorithm}
\usepackage{algpseudocode}
\usepackage{cleveref}

\usepackage[most]{tcolorbox}
\newtcolorbox[list inside=prompt,auto counter]{prompt}[1][]{
    colbacktitle=black!60,
    coltitle=white,
    boxsep=5pt,
    left=0pt,
    right=0pt,
    top=0pt,
    bottom=0pt,
    boxrule=1pt,
    #1,
}
\newcommand{\search}[1]{\textcolor{cyan}{\texttt{<search>}}#1\textcolor{cyan}{\texttt{</search>}}}
\newcommand{\info}[1]{\textcolor{brown}{\texttt{<output>}}#1\textcolor{brown}{\texttt{</output>}}}
\newcommand{\answer}[1]{\textcolor{purple}{\texttt{<answer>}}#1\textcolor{purple}{\texttt{</answer>}}}
\newcommand{\assessment}[1]{\textcolor{OliveGreen}{\texttt{<assessment>}}#1\textcolor{OliveGreen}{\texttt{</assessment>}}}

\newcommand{\method}{AdaSearch\xspace}
\newcommand{\decisionfscore}{\ensuremath{\mathbf{F1}_{\mathrm{dec}}}\xspace}

\begin{document}
\maketitle
\input{sections/0.abstract}
\input{sections/1.introduction}
\input{sections/5.related_work}
\input{sections/2.f1_metric}
\input{sections/3.method}
\input{sections/4.experiment}
\input{sections/6.conclusion}

\section*{Limitations}
Despite the advances demonstrated by our method, several limitations remain to be addressed in future research.

First, due to computational constraints, we only evaluated models up to 7B parameters. 
Future research should investigate whether our findings scale to larger architectures (e.g., 14B, 32B).

Second, because the search-decision labels are derived from the Stage-1 model's empirical solve rate, they inevitably contain sampling-based noise. Increasing the number of samples $K$ provides a more reliable estimate of the solve rate and empirically improves both \decisionfscore and EM, as shown in Figure~\ref{fig:ablation_k}. However, we do not provide a formal theoretical analysis of how this noise affects Stage-2 training, which we leave as an important direction for future work.

Third, although our approach reduces safety risks by reducing unnecessary search interactions (Appendix \ref{appendix:safety_eval}), it does not eliminate them entirely. 
Combining \method with safety alignment techniques remains an important future direction.

Finally, despite achieving superior \decisionfscore, the model still makes some errors in search invocation decisions (see Appendix \ref{case:underuse}, \ref{case:overuse}). 
Improving decision quality through fine-grained uncertainty quantification beyond binary outputs is a key area for further development.

\section*{Acknowledgments}
We sincerely thank anonymous reviewers for constructive feedback. 
This work was financially supported by the National Science and Technology Council (NSTC) and the Featured Area Research Center Program within the framework of the Higher Education Sprout Project by the Ministry of Education in Taiwan, under Grants 112-2223-E-002-012-MY5, 115-2628-E-002-023-MY4, and 115L900901. This work was also supported in part by the NVIDIA Academic Grant.

\bibliography{sections/reference}

\appendix

\input{appendix/reward_function}
\input{appendix/implementation}
\input{appendix/additional_eval}
\input{appendix/additional_analysis}
\input{appendix/algo_variant}
\FloatBarrier
\input{appendix/prompt}
\FloatBarrier
\input{appendix/case_studies}

\end{document}

%% file: sections/0.abstract.tex
\begin{abstract}
    Equipping large language models (LLMs) with search engines via reinforcement learning (RL) promises effective search agents. However, adaptively balancing internal parametric knowledge with external search remains a challenge, as overreliance on search introduces unnecessary cost and risks exposure to noisy or malicious content, while relying solely on parametric knowledge risks hallucination. 
    Prior efforts mitigate search overuse through tool-call reward shaping, which requires heavy reward engineering and conflates necessary and unnecessary search.
    To address these limitations, we revisit the evaluation of search agents through an F1-based decision metric, revealing that prior methods often overlook readily available parametric knowledge. Motivated by this, we propose \method, a simple two-stage, outcome-driven RL framework that disentangles problem-solving from the decision to search, making the decision process explicit and interpretable. Extensive experiments demonstrate that \method significantly improves search-decision quality and reduces unnecessary search calls, with only a small trade-off in QA accuracy relative to always-search.\footnote{\url{https://github.com/MiuLab/AdaSearch}}
\end{abstract}

%% file: sections/1.introduction.tex
\section{Introduction}

\input{figures/method_comparison_acl}

Large language models (LLMs) have achieved significant advances across various tasks~\citep{brown2020language,hendrycks2020measuring,team2023gemini,touvron2023llama,wei2022chain,guo2025deepseek}.
Yet they face inherent limitations: parametric knowledge alone cannot fully capture domain-specific~\citep{lewis2020retrieval,yang2024crag} or rapidly evolving information~\citep{kasai2023realtime,vu2024freshllms}, and models may hallucinate when queried beyond their knowledge boundary~\citep{ji2023survey,xu2024hallucination,huang2025survey,kalai2025language}.
To overcome these challenges, integrating LLMs with external knowledge sources has become a crucial direction~\citep{karpukhin2020dense,guu2020retrieval,nakano2021webgpt,lazaridou2022internet,shi2024replug,press2023measuring,nakano2021webgpt,feng2024knowledge}.

Early work has primarily focused on retrieval-augmented generation (RAG)~\citep{lewis2020retrieval,borgeaud2022improving}, which searches for relevant passages and appends them to the context before generation~\citep{shuster2021retrieval,ram2023context,jiang2023active,asai2024self,wei2024instructrag}.
However, the relatively static pipeline of RAG overlooks the model’s reasoning process, yielding suboptimal results~\citep{trivedi2023interleaving}.
Moreover, enabling dynamic multi-turn search is challenging due to the requirement of large-scale annotated trajectories and the non-differentiable nature of the search operation~\citep{schick2023toolformer,asai2024self,guan2025deeprag}.
To address these limitations, treating search as a tool and training tool-integrated LLMs via reinforcement learning (RL) has emerged as a promising paradigm~\citep{chen2025learning,jin2025searchr,fan2025ssrl,xiao2026meta}.
In such settings, models are trained to act as search agents, interleaving reasoning with search in a multi-turn, interactive fashion.
This allows LLMs to issue queries adaptively and integrate retrieved information into their reasoning, outperforming prior prompting-based and supervised approaches.

Nevertheless, building search agents that are truly adaptive remains challenging~\citep{jeong2024adaptive,qian2025smart,xie-etal-2026-searching}.
Ideally, a search agent should balance parametric knowledge and invoke search only when necessary. 
Excessive search calls compromise efficiency, safety, and privacy by increasing costs, exposing agents to noisy or malicious content, and risking sensitive data leaks~\citep{franklin2026ai}.
Prior attempts~\citep{huang2025reinforced,otc,song2025r1pp,song2025smart} mitigate over-search by tying rewards and evaluation to search-call counts.
However, reward shaping around search-call counts requires careful engineering (\Cref{tab:reward_design}), often involving extensive manual tuning and trial-and-error to determine suitable penalties for redundant search.
Furthermore, reward credit assignment~\citep{pignatelli2023survey} remains ambiguous: agents may superficially avoid search even when necessary to exploit rewards. Consequently, using call counts as a metric conflates necessary and redundant calls, obscuring true adaptive behavior.

Beyond efficiency, a key limitation of prior frameworks is the lack of \textit{transparency and interpretability} in the decision to invoke search. 
Since existing methods provide no direct supervision for this decision, the reasoning behind ``\textit{Should I search?}" remains implicit and difficult for users to interpret. This limitation becomes critical in high-stakes domains such as finance and medical question answering, where users must understand why an agent chooses to rely on parametric knowledge or to consult external information. 

\input{figures/qwen3b_comparison}

In this work, we introduce \textbf{Ada}ptive \textbf{Search} Agents (\textbf{\method}), a simple outcome-based RL framework that disentangles and directly optimizes two abilities:
\textit{problem solving} and
\textit{deciding whether to invoke search}.
Specifically, during training we adopt different prompts to guide LLMs in three modes:
\textit{(i)} solving the problem with parametric knowledge,
\textit{(ii)} solving the problem with search, and
\textit{(iii)} explicitly deciding whether search invocation is needed.
At inference time, the model first decides whether the question can be solved using its own knowledge via reasoning;
if so, the model answers the question directly, otherwise search calls will be invoked as part of the model's reasoning process (\Cref{fig:overview}).

Unlike prior approaches that depend on brittle reward engineering, \method simply leverages task outcomes as rewards with a training framework that provides clearer learning signals, avoids the pitfalls of ambiguous tool-count penalties, and improves search invocation decision-making through explicit reasoning (\Cref{fig:method_comparison}).
Experiments using our fine-grained search-decision F1 metric demonstrate that \method achieves the best search-decision quality across all baselines. Compared with the always-search Search-R1, \method substantially reduces unnecessary search calls (\Cref{tab:num_search}) while incurring only a small decrease in QA performance. Compared with adaptive reward-shaping baselines, it achieves both higher QA performance and better search-decision quality (\Cref{fig:qwen3b_comparison}).

In summary, our contributions are three-fold:
\vspace{-0.5em}
\begin{itemize}
    \setlength\itemsep{-0.5em}
    \item We propose \textbf{\method}, a simple yet effective RL framework that explicitly optimizes both problem solving and decision making to build adaptive search agents.
    \item \method eliminates the need for complex reward engineering, outperforms reward-shaping methods in both QA performance and search-decision F1, and reduces unnecessary search relative to Search-R1, with consistent gains across model sizes and families.
    \item \method improves interpretability and safety, enhancing the trustworthiness of agents in real-world scenarios. (Appendix~\ref{appendix:additional_eval})
\end{itemize}

%% file: figures/method_comparison_acl.tex
\begin{figure}[t!]
     \centering
     \includegraphics[width=\linewidth]{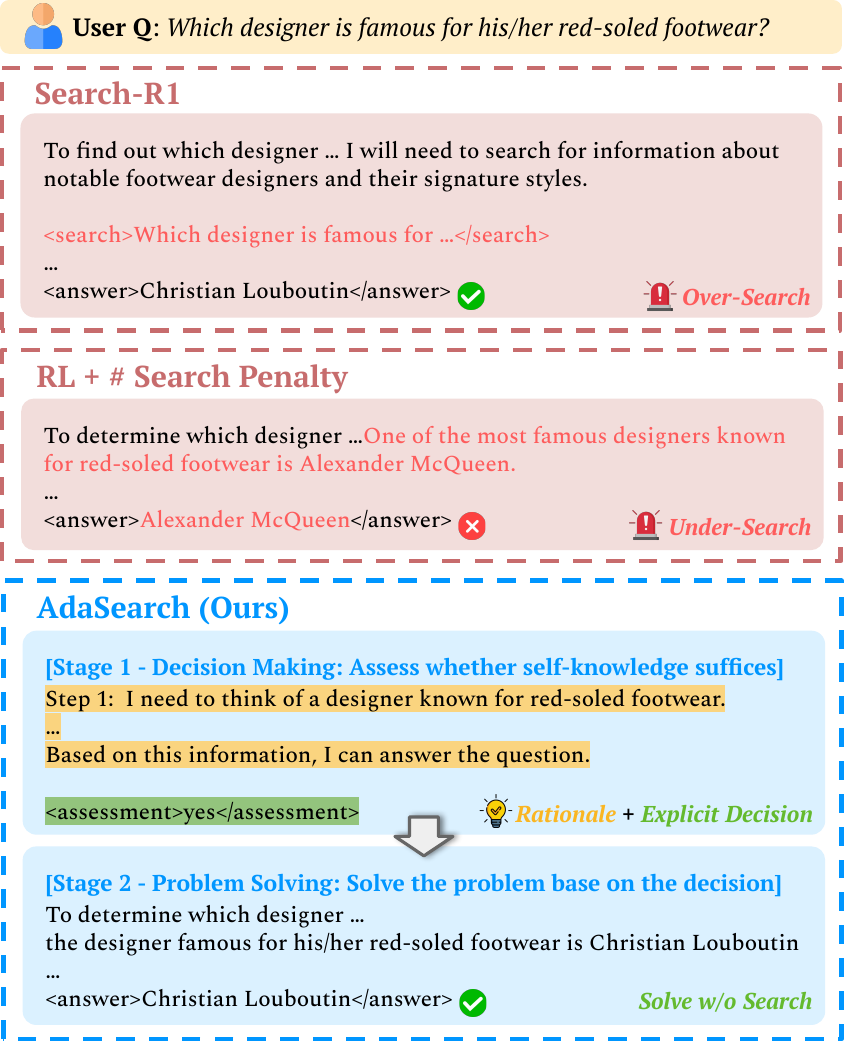}
     \caption{Comparing RL methods for search agents: \method provides transparent decisions via explicit reasoning. Search-R1 over-searches, while penalized RL under-searches; both lack explicit decision rationale.}
     \label{fig:method_comparison}
\end{figure}

%% file: figures/qwen3b_comparison.tex
\begin{figure}[htbp!]
    \centering
    \includegraphics[width=\linewidth]{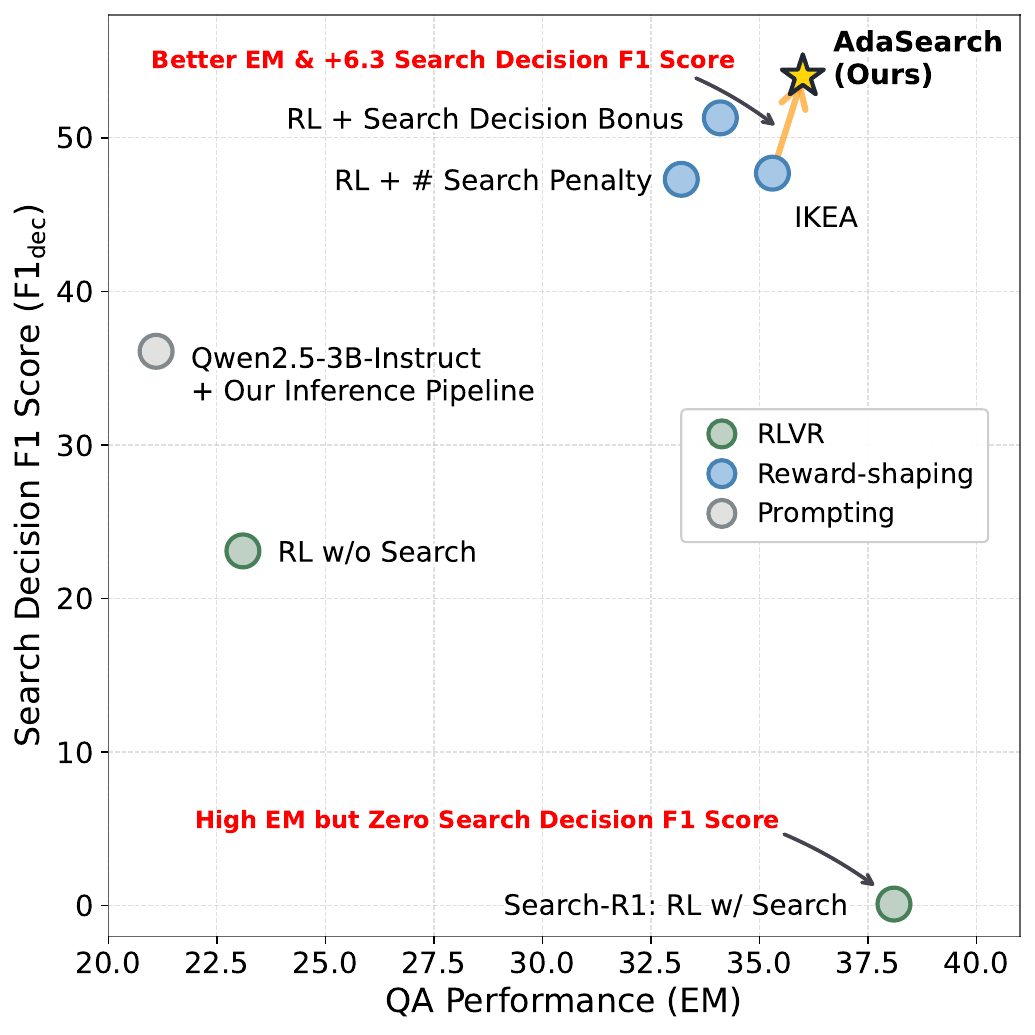}
  \caption{Among adaptive-search methods, \method achieves the highest search-decision F1 score and the best QA performance. Search-R1 achieves slightly higher EM through its always-search behavior but obtains near-zero \decisionfscore.}
  \label{fig:qwen3b_comparison}
\end{figure}

%% file: sections/5.related_work.tex
\section{Related Work}

\paragraph{Retrieval-augmented generation.}
Retrieval-augmented generation (RAG) has become a widely adopted paradigm for equipping LLMs with external knowledge~\citep{shuster2021retrieval,ram2023context,jiang2023active,asai2024self,wei2024instructrag}. 
Early approaches follow a static retrieve-and-read pipeline~\citep{lewis2020retrieval,guu2020retrieval,izacard2023atlas}, which is effective for factoid queries but relatively limited for multi-step reasoning.
Architectural variants such as RETRO~\citep{borgeaud2022improving} deliver substantial gains but require model modifications and retraining, whereas in-context RAG~\citep{ram2023context} simply prepends retrieved passages to the prompt, enabling practical off-the-shelf use.
More recent works emphasize reasoning-aware retrieval. 
IRCoT~\citep{trivedi2023interleaving} interleaves CoT~\citep{wei2022chain} with retrieval to improve multi-hop QA. 
Adaptive-RAG~\citep{jeong2024adaptive} uses an external classifier to select among no retrieval, single-step retrieval, and iterative retrieval based on query complexity.
DeepRAG~\citep{guan2025deeprag} formulates retrieval as a step-wise decision process and constructs offline RL data through a tree-search-based pipeline.
In contrast, AdaSearch targets RL-trained search agents that explicitly reason before deciding whether to invoke search.
It uses a simple generate-then-verify pipeline to identify when parametric knowledge is sufficient, enabling online RL without an external strategy classifier or tree-search-based data construction.

\paragraph{Reinforcement learning for search agents.}
RL has emerged as a powerful paradigm for developing LLM-based search agents~\cite{jin2025searchr,chen2025learning,song2025r1}.
By treating search call as tools, LLMs are trained to interleave reasoning with search in a multi-turn, interactive fashion.
Representative methods include Search-R1 and ReSearch~\citep{jin2025searchr,chen2025learning}.
Following this line of work, recent studies have focused on explicitly optimizing for search adaptivity and efficiency~\citep{otc,huang2025reinforced,song2025smart}.
Notably, OTC~\citep{otc} seeks to enhance tool productivity by penalizing search calls; IKEA~\citep{huang2025reinforced} trains agents to delineate knowledge boundaries and synergize internal and external knowledge.
However, these approaches rely on intricate reward shaping around search-call counts that requires manual tuning and suffers from ambiguous credit assignment, potentially encouraging agents to reduce search frequency without improving their search decisions.
Moreover, whether to invoke search remains an implicit decision, limiting interpretability.
Beyond search agents, uncertainty-aware methods such as R-Tuning~\citep{zhang-etal-2024-r} teaches models to refuse unknown questions through instruction tuning, while SaySelf~\citep{xu-etal-2024-sayself} uses GPT-4-generated self-reflection data for SFT before calibrating confidence through RL.
In contrast, \method learns explicit, reasoning-based search decisions directly through outcome-based RL.
At inference time, \method offers explicit decision rationales, making the agent’s behavior easier to inspect.

%% file: sections/2.f1_metric.tex
\section{Examining Search Invocation Decisions on Vanilla Search Agents}
\label{sec:self_awareness}

\subsection{Search-Decision F1 Score}

We formulate search-invocation decisions as a binary classification problem between ``search needed'' and ``search not needed.'' 
Oracle labels are obtained by evaluating whether each problem can be solved using the evaluated model’s parametric knowledge alone. 
Since our primary goal is to assess whether an agent can avoid unnecessary external interactions in such cases, we designate ``search not needed'' as the positive class. 
We report the resulting binary F1 score as the search-decision F1 score, denoted as \decisionfscore. \footnote{Because the oracle labels depend on the evaluated model's parametric knowledge, \decisionfscore should only be compared among methods post-trained from the same base model.}

Formally, we define
\begin{equation}
\decisionfscore = 
\frac{2 \cdot \text{Precision} \cdot \text{Recall}}{\text{Precision} + \text{Recall}},
\end{equation}
where precision and recall are computed by comparing the model's search decisions against oracle labels.
This metric reflects the model's capability to identify its knowledge boundaries.

Previous RL-based methods~\citep{jin2025searchr,otc,huang2025reinforced} adopt a single prompt that asks the agent to conduct the decision making process and problem solving simultaneously. 
We therefore obtain their oracle labels in two ways. 
If the model does not invoke search, we directly evaluate the correctness of its final answer. 
If the model invokes search, we generate a separate parametric-only answer using the system prompt $s_\mathrm{param}$ (Appendix~\ref{appendix:prompt}) and evaluate its correctness. 
In both cases, a correct parametric-only answer corresponds to ``search not needed'', and an incorrect answer corresponds to ``search needed.''

\subsection{Underuse of Parametric Knowledge in Vanilla Search Agents}

We evaluate prior outcome-based RL methods using Qwen2.5-7B-Base~\citep{qwen2.5}. Concretely, we include (1) Search-R1~\citep{jin2025searchr}, which equips LLMs with search access and optimizes them with outcome rewards, and (2) RL w/o Search, which conducts RL without search access and fully relies on parametric knowledge. QA performance is measured by exact match ($\mathbf{EM}$). We follow the testing setup of \citet{jin2025searchr} and report macro-averaged $\mathbf{EM}$, \decisionfscore, and confusion matrices. We re-evaluate the official Search-R1 checkpoint.

\input{tables/preliminary}

The results are summarized in Table~\ref{tab:preliminary}. For Search-R1, although search access improves $\mathbf{EM}$, its \decisionfscore is 0 because the model invokes search for every query. This overreliance, reflected in many false negatives, exemplifies tool overuse~\citep{qian2025smart}, where the model searches even when parametric knowledge suffices. In contrast, the model trained without search access achieves a higher \decisionfscore but lower $\mathbf{EM}$ due to limited internal knowledge. These results highlight the central challenge: enabling greater adaptivity by balancing parametric and external knowledge.

%% file: tables/preliminary.tex
\begin{table}[t!]
\centering
\resizebox{\linewidth}{!}{
  \setlength{\tabcolsep}{4pt}
    \begin{tabular}{lcccccc}
        \toprule 
        \textbf{Method} & \textbf{EM} & \decisionfscore & \textbf{TP} & \textbf{TN} & \textbf{FP} & \textbf{FN} \\ \midrule
        Search-R1 & 39.4 & ~~0.0 & ~~0.0 & 72.4 & ~~0.0 & 27.6 \\
        RL w/o Search & 27.6 & 27.6 & 27.6 & ~~0.0 & 72.4 & ~~0.0 \\
        \bottomrule
    \end{tabular}
}
\caption{Macro-averaged QA accuracy (EM), search decision F1 score (\decisionfscore), and confusion matrix. \textbf{TP}: true positives; \textbf{TN}: true negatives; \textbf{FP}: false positives; \textbf{FN}: false negatives.}
\label{tab:preliminary}
\end{table}

%% file: sections/3.method.tex
\section{Our Method: \method}

\input{sections/algo_adasearch}
\input{figures/overview}

We introduce \method, illustrated in Figure~\ref{fig:overview}. The key idea is to achieve adaptivity without complex reward engineering by disentangling problem solving from the decision of whether to invoke search. Separating these sub-tasks and optimizing them independently enables training policy LLM with simple outcome-based rewards. In this section, we first present the problem formulation (Sec.~\ref{subsec:problem_formulation}), and then describe the \method framework (Sec.~\ref{subsec:adasearch_framework}). Our main approach is a two-stage training framework, where the first stage targets problem solving and the second stage focuses on decision making. 
Finally, we conclude with the inference pipeline of \method (Sec.~\ref{subsec:adasearch_inference}).

\subsection{Problem Formulation}
\label{subsec:problem_formulation}

The problem setting is the same as \citet{jin2025searchr}. Given an input $x$, we let policy LLM $\pi_\theta$ interact with the search engine $\mathcal{E}$ to generate a trajectory $\tau$. Formally,
\begin{equation}
    \tau \sim \pi_\theta(\cdot \mid x; \mathcal{E}).
\end{equation}

Our goal is to train the policy $\pi_\theta$ to interact with $\mathcal{E}$ by maximizing the following RL objective:
\begin{equation}
\label{eq:rl_objective}
\begin{split}
    J(\theta) = \max_{\pi_\theta} \; & \mathbb{E}_{(x, y)\sim\mathcal{D}, \tau \sim \pi_\theta(\cdot \mid x; \mathcal{E})}\left[R(\tau, y)\right] \\
    & - \beta\,\mathbb{D}_{\mathrm{KL}} \left(\pi_\theta(\cdot \mid x; \mathcal{E}) \Vert \pi_{\mathrm{ref}}(\cdot \mid x; \mathcal{E})\right),
\end{split}
\end{equation}
where $\mathcal{D} = \{(x_i, y_i)\}^D_{i=1}$ is the training set with size $D$, $y$ is the ground truth for $x$, $R(\tau, y)$ is the reward function, and $\pi_\mathrm{ref}$ is a reference policy. We use GRPO~\citep{shao2024deepseekmath} as the RL algorithm.

\subsection{\method Training}
\label{subsec:adasearch_framework}

Our training procedure is summarized in Algorithm~\ref{algo:adasearch}. Concretely, it consists of two stages: the first stage aims to incentivize problem solving, while the second stage focuses on optimizing decision-making of search invocation.

\paragraph{Stage 1: Problem solving.}
Stage 1 focuses on incentivizing problem-solving capabilities. Concretely, we train the policy to utilize (1) parametric knowledge and (2) search call for problem solving. We design two different system prompts: (1) parametric-knowledge prompt $s_\mathrm{param}$, which requires the policy to answer using only its internal knowledge, and (2) search prompt $s_\mathrm{search}$, under which the policy
can interact with the search engine $\mathcal{E}$.
Full prompt details are provided in Appendix~\ref{appendix:prompt}.

For each $(x, y)$ in the training batch, we augment the problem $x$ with both $s_\mathrm{param}$ and $s_\mathrm{search}$, and generate two groups of rollouts: $\mathcal{R}_\mathrm{param} = \{\tau^i_\mathrm{param}\}^N_{i=1}$ and $\mathcal{R}_\mathrm{search} = \{\tau^i_\mathrm{search}\}^M_{i=1}$.
We denote this conditional generation
procedure by
$\Call{Rollout}{\pi_\theta,x,g,\mathcal{E},K_g}$,
which returns a group of $K_g$ trajectories.
When $g=\mathrm{param}$, it directly samples trajectories from
$\pi_\theta(\cdot\mid[s_\mathrm{param},x])$. When $g=\mathrm{search}$,
it samples trajectories from
$\pi_\theta(\cdot\mid[s_\mathrm{search},x];\mathcal{E})$,
allowing the access to the search engine.
We set $K_\mathrm{param}=N$ and
$K_\mathrm{search}=M$.

For each trajectory $\tau \in \mathcal{R}_\mathrm{param} \, \cup \, \mathcal{R}_\mathrm{search}$, we use regular expressions to extract the final answer $\hat{y}$ from $\tau$ and apply exact match $\mathbf{EM}(\hat{y}, y) \rightarrow \{\texttt{true}, \texttt{false}\}$, which checks whether the extracted answer is exactly the same as any of the gold answer after normalization, to check the correctness. In the following sections, we use $\mathbf{EM}$ to denote this verification procedure for simplicity.

The reward design in this stage is simply the binary correctness reward. Formally,
\begin{equation}
    \label{eq:outcome_reward}
    R(\tau, y) = \mathds{1} \left[\mathbf{EM} = \texttt{true} \right].
\end{equation}

Since $\mathcal{R}_\mathrm{param}$ and
$\mathcal{R}_\mathrm{search}$ are generated from different
prompted inputs, they constitute separate GRPO groups for
each instance. We then compute group-relative advantages
and update the policy using
GRPO~\citep{shao2024deepseekmath}.

\paragraph{Stage 2: Decision making.}
\label{subsubsec:decision_making}
Stage 2 aims to incentivize search invocation decisions. Before training, we use the stage 1 policy $\pi_{\theta_1}$ and the parametric-knowledge prompt $s_\mathrm{param}$ to generate labels. For each training instance $(x, y)$, we sample $K$ responses from $\pi_{\theta_1}$ and compute the solve rate $p$ using substring exact match ($\mathbf{SubEM}$), which checks whether any gold answer is a substring of the extracted answer after normalization:
\begin{equation}
    \label{eq:solve_rate}
    p = \frac{1}{K} \sum_{k=1}^K \mathds{1} \left[\mathbf{SubEM}(\hat{y}_k, y) = \texttt{true} \right],
\end{equation}
where $\hat{y}_k$ is the final answer extracted from the $k$-th response, and $\mathds{1}[\cdot]$ is the indicator function. We use $\mathbf{SubEM}$ instead of the stricter $\mathbf{EM}$ because the objective here is not to evaluate exact answer formatting, but to estimate whether the model can solve the problem using its parametric knowledge. $\mathbf{SubEM}$ provides a more tolerant measure that avoids penalizing semantically correct answers with minor formatting differences, making it better suited for generating decision-making supervision.

Then, we define a threshold $\rho$ and label instances with $p \geq \rho$ as solvable with parametric knowledge. We use the token \texttt{yes} as a label for such instances and the token \texttt{no} otherwise. With the generated labels, we craft the dataset $\mathcal{D}_\mathrm{decision} = \{(x_i, \ell_i)\}_{i=1}^D$, where $\ell_i \in \{\texttt{yes}, \texttt{no}\}$, for stage-2 training.

We design a decision-making system prompt $s_\mathrm{decision}$ that requires explicit reasoning before producing the final decision $\ell$, as detailed in Table~\ref{prompt:decision}. Such explicit reasoning increases transparency and interpretability of the decision-making process, making it clearer why the agent chooses to invoke search or rely on self-knowledge. The reward function in this stage remains a simple binary outcome reward, identical to that used in stage 1 (Eq.~\ref{eq:outcome_reward}).

\subsection{\method Inference}
\label{subsec:adasearch_inference}

We adopt a two-stage inference pipeline (Fig.~\ref{fig:overview}). 
In stage 1, the model is prompted with the system instruction $s_\mathrm{decision}$ to decide whether it can answer the query using self-knowledge through explicit reasoning. 
In stage 2, the model receives the corresponding problem-solving instruction: if it decides it can answer with self-knowledge, use $s_\mathrm{param}$; otherwise, use $s_\mathrm{search}$. 
Note that stage-1 history is not prepended in stage 2.

%% file: sections/algo_adasearch.tex
\begin{algorithm}[t!]
\small
\caption{\method Training}
\label{algo:adasearch}
\begin{algorithmic}[1]
\Require Base policy $\pi_{\theta_b}$, search engine $\mathcal{E}$,
training set $\mathcal{D}$, prompts
$s_{\mathrm{param}},s_{\mathrm{search}},s_{\mathrm{decision}}$,
rollout sizes $N,M$, stage-2 labeling hyperparameters $K$ and $\rho$

\State $\pi_\theta \gets \pi_{\theta_b}$
\State $K_{\mathrm{param}}\gets N$;
       $K_{\mathrm{search}}\gets M$

\Statex \textit{\textcolor{blue}{/* Stage 1: Problem-solving training */}}
\For{$t=1,\ldots,T$}
    \State Sample a batch $\mathcal{B}\sim\mathcal{D}$
    \State $\mathcal{R}\gets\emptyset$

    \For{$(x_i,y_i)\in\mathcal{B}$}
        \For{$g\in\{\mathrm{param},\mathrm{search}\}$}
            \State $\mathcal{R}_g^i
            \gets
            \Call{Rollout}
            {\pi_\theta,x_i,g,\mathcal{E},K_g}$
            \State Compute rewards $\mathbf{r}_g^i$
            for $\mathcal{R}_g^i$ using
            Eq.~\ref{eq:outcome_reward}
            \State $\mathcal{R}\gets\mathcal{R}\cup\mathcal{R}_g^i$
        \EndFor
    \EndFor

    \State Update $\pi_\theta$ with GRPO using the rewarded rollout groups in $\mathcal{R}$
\EndFor
\State $\pi_{\theta_1}\gets\pi_\theta$

\Statex \textit{\textcolor{blue}{/* Stage 2: Decision-making training */}}
\State Generate $\mathcal{D}_{\mathrm{decision}}$ using
$\pi_{\theta_1}$, $\mathcal{D}$, $s_{\mathrm{param}}$,
$K$, and $\rho$
\State Update $\pi_\theta$ on
$\mathcal{D}_{\mathrm{decision}}$ with GRPO under
$s_{\mathrm{decision}}$
\State \Return $\pi_\theta$
\end{algorithmic}
\end{algorithm}

%% file: figures/overview.tex
\begin{figure*}[t!]
    \centering
    \includegraphics[width=\textwidth]{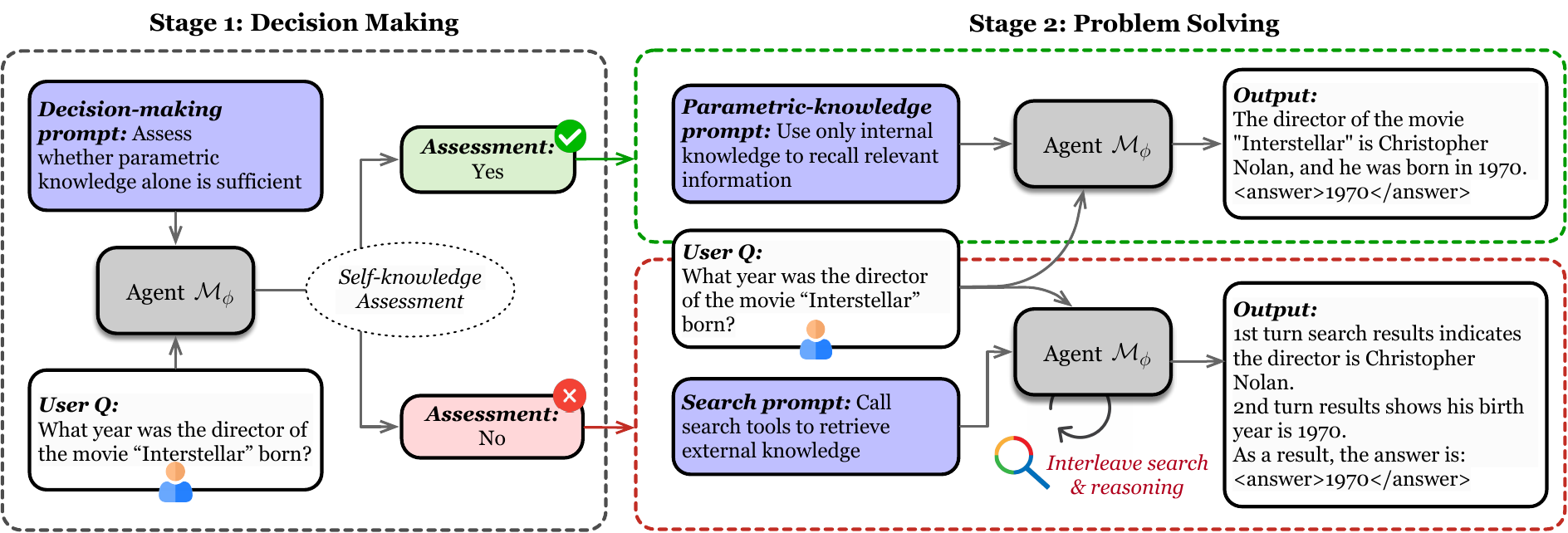}
    \caption{Overview of \method during inference. In stage 1, the agent explicitly reasons to decide whether the query can be solved via parametric knowledge. In stage 2, it follows the parametric-knowledge prompt if sufficient; otherwise, it switches to the search prompt to interleave reasoning and search for the final answer.}
    \label{fig:overview}
\end{figure*}

%% file: sections/4.experiment.tex
\section{Experiments}

\input{tables/main}

\subsection{Experiment Setup}
\label{subsec:exp_setup}

\paragraph{Models.}
We conduct experiments on Qwen2.5~\citep{qwen2.5} and Llama-3.2~\citep{grattafiori2024llama}. Due to compute constraints, our main experiments use the 3B Instruct variants, with scaling results presented in Section~\ref{par:model_scale}.

\paragraph{Search environment.}
Following \citet{jin2025searchr}, we use E5~\citep{wang2022text} as the retriever over a 2018 Wikipedia dump, with the implementation from \citet{griggs2025skrylv01}. For each query, we return the top-3 documents and limit the number of search calls to at most three. Additional results using different retrievers are provided in Appendix~\ref{appendix:bm25_results}.

\paragraph{Baselines.} 
We consider three categories of baselines. For \textbf{prompting baselines}, we select Direct inference, Chain-of-Thought (CoT) Reasoning~\citep{wei2022chain}, and RAG~\citep{lewis2020retrieval}. For \textbf{outcome-RL baselines}, we apply GRPO~\citep{shao2024deepseekmath} to train models on the parametric prompt $s_\mathrm{param}$ (RL w/o search) and on the search prompt $s_\mathrm{search}$ (Search-R1). For \textbf{reward-shaping baselines}, we include IKEA (\citet{huang2025reinforced}, Table~\ref{tab:reward_design}) as a representative method, and also design two baselines. The first one is denoted as ``Naive Shaping". Its reward function is defined as
\begin{equation}
\label{eq:naive_shaping}
    R(\tau, y) = \begin{cases}
        1.0 - \lambda \cdot (\text{\# search}) & \text{if } \mathbf{EM} = \texttt{true}, \\
        0 & \text{otherwise,}
    \end{cases}
\end{equation}
which is conceptually identical to OTC and R1-Searcher++~\citep{otc,song2025r1pp} but easier to implement, assigning higher rewards to correct rollouts with fewer search calls. We set $\lambda = 0.05$.

In addition, we introduce another reward-shaping baseline designed to enforce prompt-level search decisions, which closely resembles the objective of \method. We refer to it as ``Decision Shaping". Its reward function is defined as
\begin{equation}
\label{eq:decision_shaping}
\begin{split}
    R(\tau, y) = & \mathds{1}\left[\mathbf{EM} = \texttt{true}\right] \\
    & + \alpha \cdot \begin{cases}
        \mathds{1} \left[ \tau \text{ has search} \right] & \text{if } p < \rho, \\
        \mathds{1} \left[ \tau \text{ has no search} \right] & \text{otherwise},
    \end{cases}
\end{split}
\end{equation}
where $p$ is the empirical solve rate defined in Eq~\ref{eq:solve_rate}, $\rho$ is the threshold for $p$, and $\alpha$ is the hyperparameter controlling the bonus for correct decisions. This baseline rewards the model for making appropriate search decisions. We estimate $p$ for each sample using the base policy before RL training and set both $\alpha$ and $\rho$ to 0.5. Finally, the prompts used for each baseline and the hyperparameter analysis are detailed in Appendix~\ref{appendix:prompt} and~\ref{appendix:reward_shaping_hyperparam}, respectively.

\paragraph{Training.}
To ensure fair comparison, we construct a difficulty-balanced training set of 8,192 samples following the procedure of \citet{huang2025reinforced}. A validation set is constructed using the same procedure and used for checkpoint selection. Full details for dataset construction are presented in Appendix~\ref{appendix:dataset_construction}. For \method stage 2, we use the same threshold $\rho = 0.5$. Full training details are provided in Appendix~\ref{appendix:training_details}.

\paragraph{Evaluation.} 
Following \citet{jin2025searchr}, we use $\mathbf{EM}$ as the task metric, apply greedy decoding, and evaluate on standard QA benchmarks.
For single-hop QA, we use Natural Questions~\citep{natural_questions}, TriviaQA~\citep{joshi-etal-2017-triviaqa}, and PopQA~\citep{mallen-etal-2023-trust}.
For multi-hop QA, we use HotpotQA~\citep{yang-etal-2018-hotpotqa}, 2WikiMultihopQA~\citep{ho-etal-2020-constructing}, MuSiQue~\citep{trivedi-etal-2022-musique}, and Bamboogle~\citep{press2023measuring}.
We also report \decisionfscore (Section~\ref{sec:self_awareness}) to measure the quality of search invocation decisions. 
Interpretability, safety, and universally unanswerable question evaluations are provided in Appendix~\ref{appendix:additional_eval}. Balanced accuracy~\citep{5597285} for search decisions and multi-seed inference results are reported in Appendices~\ref{appendix:balanced_acc} and~\ref{appendix:multi_seed_results}, respectively.

\subsection{Main Results}

\paragraph{Comparison against prompting and outcome-RL baselines.}
Table~\ref{tab:main} summarizes the results of prompting and outcome-RL baselines. In terms of task performance $\mathbf{EM}$, Search-R1 achieves the best overall results, and the RAG baseline consistently outperforms all prompting-based methods. This pattern reflects the characteristics of these benchmarks, where retrieval generally offers an advantage over purely parametric reasoning; the performance gap between RL w/o Search and Search-R1 further supports this observation.
Regarding search decision, however, both RAG and Search-R1 obtain scores close to zero across benchmarks due to their always-search behavior. The results of Search-R1 highlight the deficiency of naive QA outcome rewards in balancing parametric and external knowledge. Non-search baselines achieve higher \decisionfscore, but because they never explicitly assess their own knowledge boundaries, their decision-making remains suboptimal.
In comparison, our \method maintains competitive $\mathbf{EM}$ while substantially improving search invocation decision-making across model families, achieving 54\%\ to 60\% relative gain in \decisionfscore over Search-R1. 
These results highlight that our method can effectively elicit better decisions while incurring only a small $\mathbf{EM}$ decrease relative to Search-R1.

\paragraph{Comparison against reward-shaping baselines.}
As shown in Table~\ref{tab:main}, our \method consistently outperforms intricate reward-shaping baselines in both $\mathbf{EM}$ and \decisionfscore. First, Naive Shaping generally underperforms other baselines in $\mathbf{EM}$, suggesting that naively penalizing tool usage might hurt performance. 
On the other hand, although Decision Shaping generally shows improvements in both $\mathbf{EM}$ and \decisionfscore over Naive Shaping, the results are still suboptimal. We suspect that this may be due to distribution shifts, where solve rates $p$ change during training. Consequently, using initial offline-generated labels introduces label noise.
Lastly, IKEA~\citep{huang2025reinforced} generally outperforms the other reward-shaping baselines in both $\mathbf{EM}$ and \decisionfscore (except on Qwen2.5-3B-Instruct). Its reward design can be viewed as combining both Naive Shaping (e.g., penalizing tool usage on correct trajectories) and Decision Shaping (e.g., adding bonuses on incorrect trajectories that involve search calls).
However, \method still consistently outperforms all reward-shaping baselines across models, especially on challenging multi-hop QA where search is more genuinely required. This suggests that reward shaping often over-optimizes for reducing tool calls and fails to adapt to query difficulty. In contrast, \method strengthens search invocation decision-making while achieving better task performance through explicit binary supervision.

\input{figures/analysis_self_awareness_acl}
\input{tables/method_variant_comparison}

\subsection{Analysis}

\paragraph{\method achieves better search invocation decision-making.}

\Cref{fig:confusion_matrix}  and \Cref{tab:em_f1_prec_recall} presents the confusion matrices and precision–recall–F1 scores across different methods and models. For Naive Shaping, we observe the lowest false negative rates but substantially higher false positive rates for both models (Fig.~\ref{fig:confusion_matrix}). This pattern suggests that naively penalizing search usage causes the model to underuse search: it spuriously avoids invoking search even when its parametric knowledge is insufficient. In contrast, \method achieves comparable true positive and false negative rates while significantly reducing the false positive rate.

For the other reward-shaping baselines, although they show more balanced false positive and false negative rates, \method still outperforms them in both \decisionfscore and true positive rate. A closer examination shows that \method attains precision on par with Decision Shaping and IKEA, while achieving noticeably higher recall (Fig.~\ref{tab:em_f1_prec_recall}).

\paragraph{Two-stage training vs joint optimization.}

To validate the benefit of two-stage training over joint optimization, we implement an end-to-end variant, \textbf{\method-E2E}, which jointly optimizes problem solving and search invocation decision-making. This design enables on-the-fly label generation for search decisions by using the empirical solve rate of parametric knowledge at each step to produce labels. Further details are provided in Appendix~\ref{appendix:adasearch_e2e}.

As shown in Figure~\ref{tab:em_f1_prec_recall}. \method-E2E consistently underperforms \method, suggesting that the two-stage approach is more effective.

\paragraph{RL vs SFT in stage-2 training.}
A common approach for teaching models to express uncertainty is supervised fine-tuning (SFT)~\citep{lin2022teaching,zhang-etal-2024-r}. In stage 2, we implement an SFT variant, \textbf{\method-SFT}. After computing solve rates (Sec.~\ref{subsubsec:decision_making}), we directly construct target responses: ``\assessment{yes}" when $p \ge 0.5$, and ``\assessment{no}" otherwise. Details are provided in Appendix~\ref{appendix:implementation}.

The results are shown in Table~\ref{tab:em_f1_prec_recall}. While \method-SFT improves \decisionfscore over Search-R1 and achieves slightly higher $\mathbf{EM}$ on Qwen, it still underperforms \method on Llama across both metrics. 
The larger \decisionfscore gap on MuSiQue and Bamboogle highlights that reasoning-before-assessment is crucial for OOD generalization.

\input{figures/learning_curves}

\paragraph{Learning curves.} 
The learning curves are shown in Figure~\ref{fig:learning_curves}, with the training reward trends provided in Appendix~\ref{appendix:training_rewards}. In stage 1, the model’s performance on both search-based and parametric problem solving matches the single-task baselines, Search-R1 and RL w/o Search (Fig.~\ref{fig:ablation_problem_solving}). In stage 2, the quality of search invocation decisions improves substantially and task performance continues to rise (Figure~\ref{fig:ablation_performance}). Moreover, Stage 2 does not degrade problem-solving ability (Fig.~\ref{fig:ablation_problem_solving}), consistent with findings that on-policy RL mitigates forgetting~\citep{shenfeld2025rl,chen2025retaining}.

\paragraph{Stage-2 labeling hyperparameters.}
To understand the effect of labeling hyperparameters for stage-2 training, we conduct ablation studies on (1) the number of responses $K$ used to estimate the empirical solve rate and (2) the solve-rate threshold $\rho$ for stage-2 labeling. Additional analysis is provided in Appendix~\ref{appendix:stage2_ablation}.

For (1) (Fig.~\ref{fig:ablation_k}), increasing $K$ improves both \decisionfscore and $\mathbf{EM}$ via more reliable solve-rate estimates. Notably, even $K=1$ performs well (35.1\% $\mathbf{EM}$ and 52.1\% \decisionfscore), likely because RL sharpens output distributions~\citep{yue2025does,he-etal-2025-rewarding}, allowing a single sample to produce effective signals. 
For (2) (Fig.~\ref{fig:ablation_p}), $\rho$ has limited effect when $\rho \le 0.5$, aside from a drop in $\mathbf{EM}$ at $\rho = 0.3$. 
As $\rho$ increases further, \decisionfscore declines while $\mathbf{EM}$ rises, since higher $\rho$ encourages more \texttt{no} assessments, increasing search usage.


\input{figures/ablation}

\paragraph{Model sizes.}
\label{par:model_scale}
To assess the robustness of our method across model scales, we train Qwen2.5-7B-Instruct and compare \method with Search-R1 and Naive Shaping. As shown in Figure~\ref{fig:qwen7b_results}, \method boosts \decisionfscore by roughly 60\% compared to Search-R1. Relative to Naive Shaping, \method improves both $\mathbf{EM}$ (+2.1\%) and \decisionfscore (+4.2\%). These results indicate that \method generalizes well to larger model scales.

\paragraph{Average number of searches.}
\label{par:avg_search}
We report the average search calls in \Cref{tab:num_search}. \method reduces unnecessary search usage by 34–38\% compared to Search-R1, and its frequency is similar to Decision Shaping since both rely on self-knowledge for prompt-level decision making. \method uses slightly more searches than Naive Shaping and IKEA, which is expected because it does not directly optimize for minimizing tool calls. Further analysis of efficiency is provided in Appendix~\ref{appendix:efficiency}.

%% file: tables/main.tex
\definecolor{lightblue}{RGB}{224,242,254}

\begin{table*}[t!]
  \centering
  \resizebox{\linewidth}{!}{
  \setlength{\tabcolsep}{4pt} 
    \begin{tabular}{lcccccccccccccccc}
      \toprule
      \multirow{3}{*}{\bf Method} &
      \multicolumn{6}{c}{\bf General QA} &
      \multicolumn{8}{c}{\bf Multi-Hop QA} &
      \multicolumn{2}{c}{\multirow{2}{*}{\bf Avg.}} \\
      \cmidrule(lr){2-7}\cmidrule(lr){8-15}
      & \multicolumn{2}{c}{\bf NQ} 
      & \multicolumn{2}{c}{\bf TQ} 
      & \multicolumn{2}{c}{\bf PopQA} 
      & \multicolumn{2}{c}{\bf HotpotQA}
      & \multicolumn{2}{c}{\bf 2Wiki}
      & \multicolumn{2}{c}{\bf MuSiQue}
      & \multicolumn{2}{c}{\bf Bamboogle}
      & & \\ 
      \cmidrule(lr){2-3}\cmidrule(lr){4-5}\cmidrule(lr){6-7}
      \cmidrule(lr){8-9}\cmidrule(lr){10-11}\cmidrule(lr){12-13}\cmidrule(lr){14-15}\cmidrule(lr){16-17}
      & \bf EM & \decisionfscore & \bf EM & \decisionfscore & \bf EM & \decisionfscore & \bf EM & \decisionfscore & \bf EM & \decisionfscore & \bf EM & \decisionfscore & \bf EM & \decisionfscore & \bf EM & \decisionfscore \\
      \midrule
      \multicolumn{17}{l}{\bf Qwen2.5-3B-Instruct} \\
      \multicolumn{17}{l}{\textit{Prompting Baselines}} \\
      \quad Direct Answer & ~~8.1 & ~~8.1 & 24.1 & 24.1 & ~~7.8 & ~~7.8 & 13.2 & 13.2 & 19.5 & 19.5 & ~~1.4 & ~~1.4 & ~~5.6 & ~~5.6 & 11.4 & 11.4 \\
      \quad CoT & 14.1 & 14.1 & 38.2 & 38.2 & 13.5 & 13.5 & 17.3 & 17.3 & 22.8 & 22.8 & ~~3.9 & ~~3.9 & 24.0 & 24.0 & 19.1 & 19.1 \\
      \quad RAG & 31.7 & ~~0.0 & 52.7 & ~~0.0 & 38.1 & ~~0.0 & 24.6 & ~~0.0 & 20.1 & ~~0.0 & ~~4.4 & ~~0.0 & ~~8.0 & ~~0.0 & 25.7 & ~~0.0 \\
      \midrule
      \multicolumn{17}{l}{\textit{Outcome RL Baselines}} \\
      \quad RL w/o search & 21.7 & 21.7 & 44.9 & 44.9 & 17.3 & 17.3 & 20.5 & 20.5 & 28.0 & 28.0 & ~~5.9 & ~~5.9 & 23.2 & 23.2 & 23.1 & 23.1 \\
      \quad Search-R1 & \bf 42.7 & ~~0.0 & \bf 60.1 & ~~0.3 & \underline{43.9} & ~~0.0 & \bf 36.9 & ~~0.1 & \underline{37.3} & ~~0.2 & \bf 16.1 & ~~0.0 & \bf 29.6 & ~~0.0 & \bf 38.1 & ~~0.1 \\
      \midrule
      \multicolumn{17}{l}{\textit{Reward-Shaping Baselines}} \\
      \quad Naive Shaping & 37.8 & 41.6 & 54.8 & 66.9 & 42.2 & 55.3 & 29.8 & 49.3 & 33.1 & 58.4 & 9.9 & 15.8 & 24.8 & 43.8 & 33.2 & 47.3\\
      \quad Decision Shaping & 37.7 & \underline{45.9} & 56.1 & \underline{70.3} & 41.8 & \bf 61.3 & 32.1 & \underline{57.5} & 35.9 & \underline{61.1} & 12.5 & \underline{21.7} & 22.4 & 41.1 & 34.1 & \underline{51.3} \\
      \quad IKEA & \underline{41.4} & 37.7 & \underline{59.5} & 62.4 & \bf 44.2 & 55.2 & 32.0 & \bf 58.1 & 35.0 & 60.2 & 11.5 & 14.2 & 23.2 & \underline{46.0} & 35.3 & 47.7 \\
      \midrule
      \rowcolor{lightblue!100}
      \textbf{\method (Ours)} & 37.9 & \bf 49.0 & 56.8 & \bf 71.7 & 42.8 & \underline{58.8} & \underline{33.4} & \underline{57.5} & \bf 38.5 & \bf 65.5 & \underline{14.4} & \bf 25.2 & \underline{28.0} & \bf 50.0 & \underline{36.0} & \bf 54.0 \\
      \midrule
      \multicolumn{17}{l}{\bf Llama-3.2-3B-Instruct} \\
      \multicolumn{17}{l}{\textit{Prompting Baselines}} \\
      \quad Direct Answer & 18.8 & 18.8 & 42.3 & 42.3 & 16.4 & 16.4 & 12.5 & 12.5 & 16.2 & 16.2 & ~~3.0 & ~~3.0 & ~~8.8 & ~~8.8 & 16.9 & 16.9 \\
      \quad CoT & 25.0 & 25.0 & 45.6 & 45.6 & 15.8 & 15.8 & 16.3 & 16.3 & 11.3 & 11.3 & ~~4.5 & ~~4.5 & 35.2 & 35.2 & 22.0 & 22.0 \\
      \quad RAG & 29.4 & ~~0.0 & 53.0 & ~~0.0 & 35.7 & ~~0.0 & 20.6 & ~~0.0 & ~~8.3 & ~~0.0 & ~~4.1 & ~~0.0 & 12.8 & ~~0.0 & 23.4 & ~~0.0 \\
      \midrule
      \multicolumn{17}{l}{\textit{Outcome RL Baselines}} \\
      \quad RL w/o search & 38.2 & 38.2 & 54.6 & 54.6 & 23.1 & 23.1 & 23.7 & 23.7 & 26.7 & 26.7 & ~~6.7 & ~~6.7 & 34.4 & 34.4 & 29.6 & 29.6 \\
      \quad Search-R1 & \bf 46.1 & ~~1.7 & \bf 64.4 & ~~2.5 & \bf 44.8 & ~~1.5 & \bf 38.2 & ~~6.6 & \underline{34.8} & ~~1.0 & \bf 14.6 & ~~1.3 & \bf 42.4 & ~~0.0 & \bf 40.7 & ~~2.1 \\
      \midrule
      \multicolumn{17}{l}{\textit{Reward-Shaping Baselines}} \\
      \quad Naive Shaping & 41.2 & 57.6 & 59.7 & 75.3 & 39.3 & 56.3 & 31.2 & 52.6 & 33.5 & \underline{54.8} & 11.5 & 21.1 & 36.8 & 55.9 & 36.2 & 53.4 \\
      \quad Decision Shaping & 41.4 & \underline{58.8} & 61.5 & 75.4 & \underline{41.2} & \underline{61.4} & 33.7 & 57.5 & 33.7 & 48.1 & 13.0 & 31.4 & 32.0 & 58.1 & 36.6 & 55.8 \\
      \quad IKEA & 42.2 & 58.6 & 62.0 & \underline{76.2} & 41.1 & 59.9 & 32.3 & \underline{58.2} & 32.9 & 51.3 & 11.2 & \underline{31.9} & \underline{40.8} & \underline{63.1} & 37.5 & \underline{57.0} \\
      \midrule
      \rowcolor{lightblue!100}
      \textbf{\method (Ours)} & \underline{43.5} & \bf 62.7 & \underline{62.9} & \bf 79.2 & 40.4 & \bf 62.3 & \underline{34.4} & \bf 59.2 & \bf 36.2 & \bf 60.1 & \underline{13.2} & \bf 36.5 & \bf 42.4 & \bf 64.2 & \underline{39.0} & \bf 60.6 \\
      \bottomrule
    \end{tabular}}
  \caption{Results of different model across benchmarks. $\mathbf{EM}$ denotes exact match. \decisionfscore denotes the search invocation decision-making score (see Sec.~\ref{sec:self_awareness}). \textbf{Avg} is computed by averaging metrics over benchmarks. The best score is shown in \textbf{bold} and the second-best in \underline{underlined}.}
  \label{tab:main}
\end{table*}

%% file: figures/analysis_self_awareness_acl.tex
\begin{figure}[t!]
    \centering
    \includegraphics[width=\linewidth]{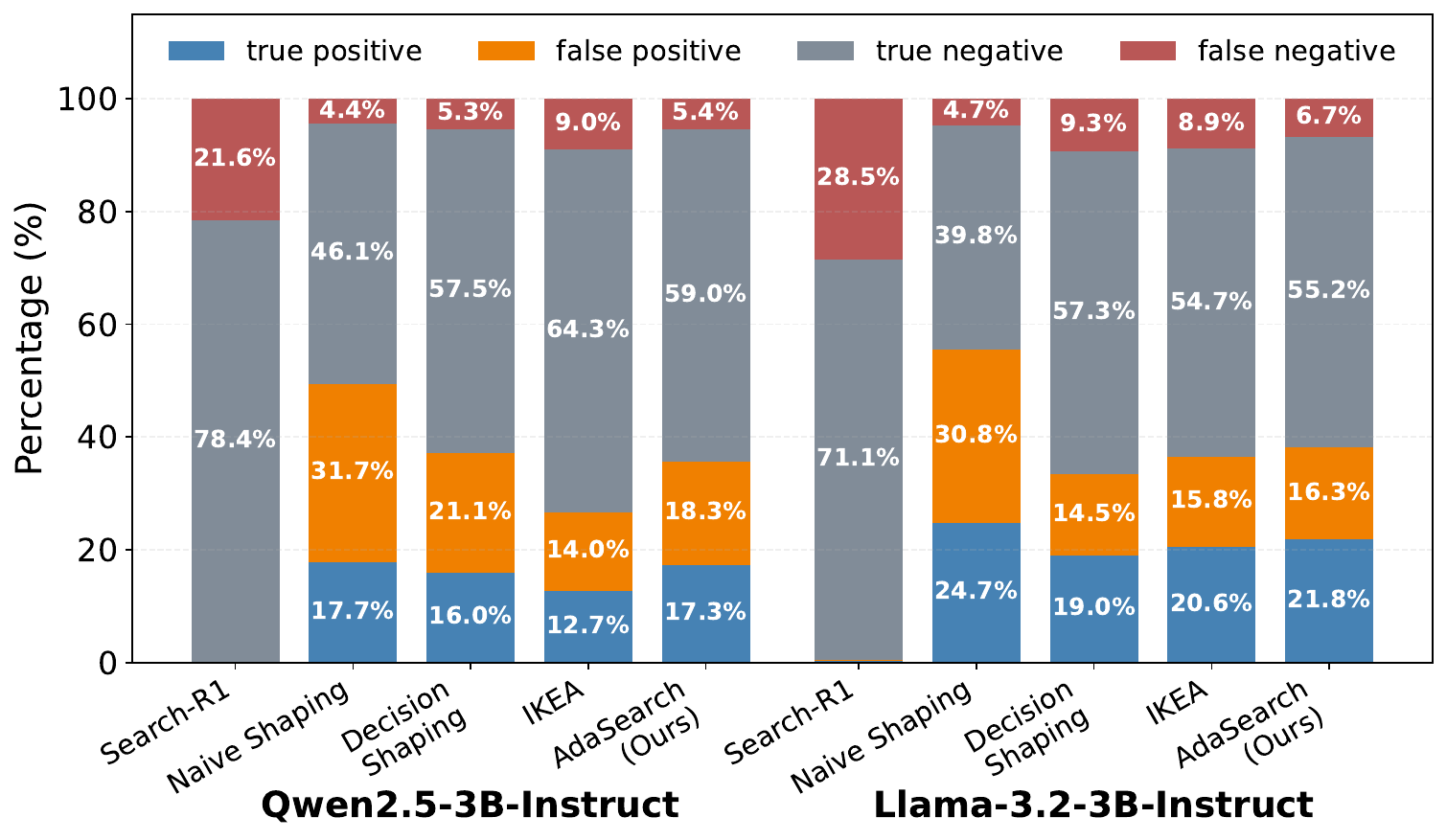}
     \caption{Confusion matrix of different RL methods.}
     \label{fig:confusion_matrix}
\end{figure}

%% file: tables/method_variant_comparison.tex
\begin{table}[t!]
    \centering
    \resizebox{\linewidth}{!}{
        \begin{tabular}{llcccc}
            \toprule
            \textbf{Model} & \textbf{Method} & $\mathbf{EM}$ & \decisionfscore & \textbf{Prec.} & \textbf{Rec.} \\
            \midrule
            \multirow{7}{*}{\begin{tabular}[c]{@{}l@{}} \bf Qwen2.5-\\ \bf 3B-Instruct \end{tabular}}
            & Search-R1 & \bf 38.1 & 0.1 & 39.0 & 0.0 \\
            & Naive Shaping & 33.2 & 47.3 & 36.1 & \bf 76.8 \\
            & Decision Shaping & 34.1 & 51.3 & 42.0 & 70.3 \\
            & IKEA & 35.3 & 47.6 & 45.0 & 53.4 \\
            \cdashline{2-5}
            & \method & 36.0 & \bf 54.0 & 45.1 & 68.8 \\
            & \method-E2E & 35.2 & 51.3 & 44.1 & 63.9 \\
            & \method-SFT & 37.0 & 51.1 & \bf 45.9 & 59.5 \\
            \midrule
            \multirow{7}{*}{\begin{tabular}[c]{@{}l@{}} \bf Llama-3.2-\\ \bf 3B-Instruct \end{tabular}}
            & Search-R1      & \bf 40.7 & ~~2.1 & 47.3 & ~~1.1 \\
            & Naive Shaping  & 36.2 & 53.4 & 41.4 & \bf 79.0 \\
            & Decision Shaping & 36.6 & 55.8 & 52.1 & 60.5 \\
            & IKEA           & 37.5 & 57.0 & 51.4 & 64.5 \\ 
            \cdashline{2-5}
            & \method        & 39.0 & \bf 60.6 & \bf 53.0 & 71.4 \\
            & \method-E2E    & 37.7 & 58.6 & 48.1 & 75.9 \\
            & \method-SFT    & 36.6 & 55.6 & 46.6 & 69.4 \\
            \bottomrule
        \end{tabular}
    }
    \caption{Averaged $\mathbf{EM}$, \decisionfscore, precision, and recall.}
    \label{tab:em_f1_prec_recall}
\end{table}

%% file: figures/learning_curves.tex
\begin{figure}[t!]
     \centering
     \begin{subfigure}{0.48\linewidth}
         \centering
         \includegraphics[width=\textwidth]{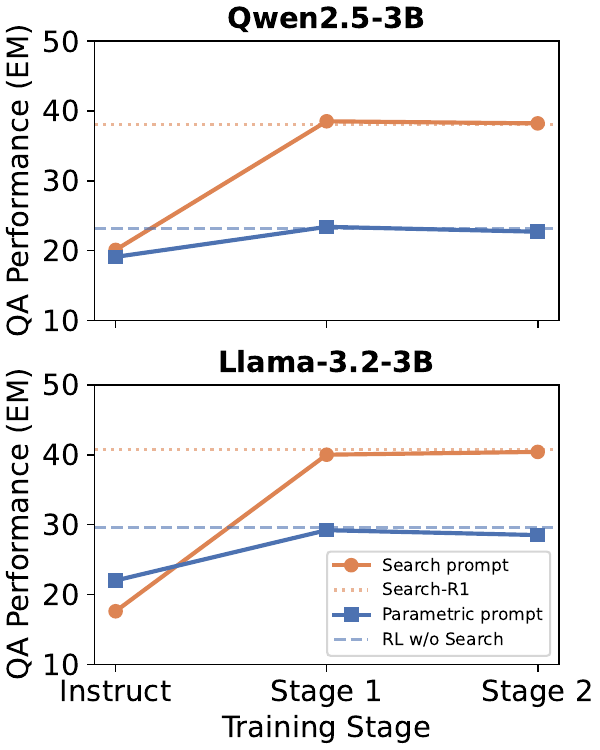}
         \caption{Problem-solving $\mathbf{EM}$.}
         \label{fig:ablation_problem_solving}
     \end{subfigure}
     \begin{subfigure}{0.48\linewidth}
         \centering
         \includegraphics[width=\textwidth]{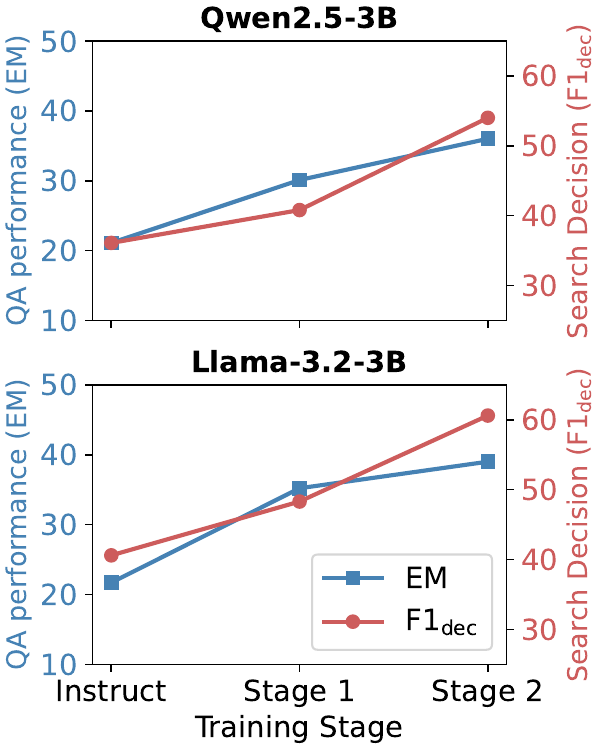}
         \caption{Avg. $\mathbf{EM}$ and \decisionfscore.}
         \label{fig:ablation_performance}
     \end{subfigure}

     
     
     \caption{Averaged performance. (a) Stage 1 substantially boosts both parametric and search-based problem solving, with Stage 2 maintaining this performance. (b) Both test $\mathbf{EM}$ and search decision-making (\decisionfscore) improve steadily during training.}
    \label{fig:learning_curves}
\end{figure}

%% file: figures/ablation.tex
\begin{figure}[t!]
     \centering
     \begin{subfigure}[htbp!]{0.235\textwidth}
         \centering
         \includegraphics[width=\textwidth]{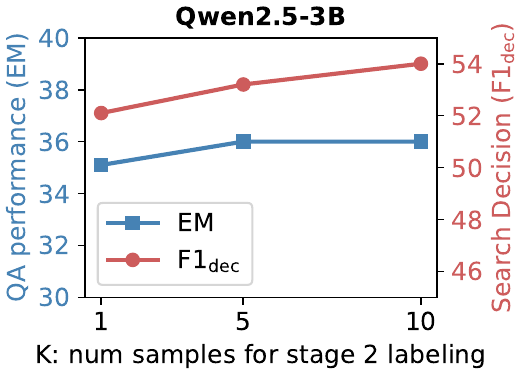}
         \caption{Number of samples $K$.}
         \label{fig:ablation_k}
     \end{subfigure}
     \hfill
     \begin{subfigure}[htbp!]{0.235\textwidth}
         \centering
         \includegraphics[width=\textwidth]{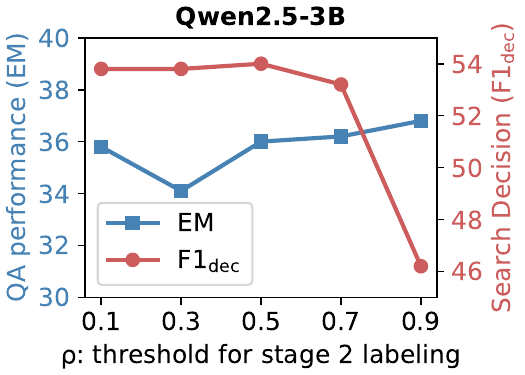}
         \caption{Solve rate threshold $\rho$.}
         \label{fig:ablation_p}
     \end{subfigure}
    \begin{subfigure}[htbp!]{0.235\textwidth}
        \centering
        \includegraphics[width=0.95\textwidth]{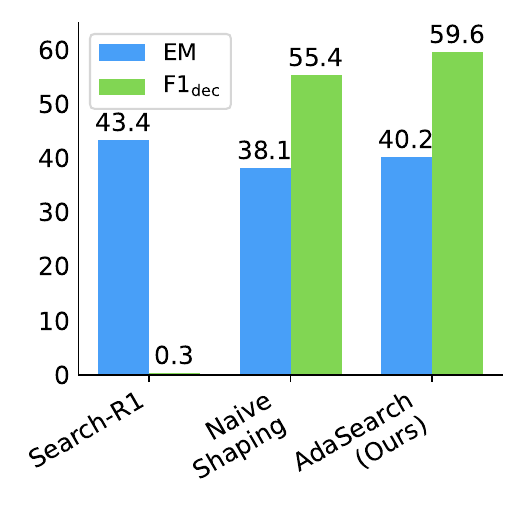}
        \vspace{-1em}
         \caption{Model size (7B).}
         \label{fig:qwen7b_results}
     \end{subfigure}
     \hfill
     \begin{subfigure}[htbp!]{0.235\textwidth}
        \centering
        \vspace{1em}
        \resizebox{\textwidth}{!}{
        \begin{tabular}{lcc}
            \toprule
             \textbf{Method} & \textbf{Qwen} & \textbf{Llama} \\
             \midrule
             Search-R1 & 1.64 & 1.75  \\
             Naive & 0.77 & 0.69 \\
             Decision & 1.00 & 1.11 \\
             IKEA & 0.88 & 0.79 \\
             \hdashline
             \method & 1.08 & 1.10 \\
             \bottomrule
        \end{tabular}
        }
        \vspace{0.9em}
        \caption{Avg. number of search.}
        \label{tab:num_search}
     \end{subfigure}
     \vspace{-0.5em}
     \caption{Ablations of \method.}
     \label{fig:ablations}
\end{figure}

%% file: sections/6.conclusion.tex
\section{Conclusion}

In this work, we analyze existing search agents and find that they struggle to recognize the limits of their parametric knowledge, leading to excessive search. 
Motivated by this observation, we propose \method, a simple two-stage, outcome-driven RL framework that improves search invocation decisions without complex reward engineering. 
Across different models, \method makes the most reliable search decisions, reduces unnecessary search calls, and provides more transparent and interpretable rationales for when to search, while incurring only minimal drops in QA accuracy. 
These results demonstrate the effectiveness and generalizability of \method for building adaptive and trustworthy search agents.

%% file: appendix/reward_function.tex
\section{Comparison on Reward Functions of Different RL Methods}

We summarize the reward designs of different methods in \Cref{tab:reward_design}. \textbf{OTC}~\citep{otc} is designed to minimize search calls and maximize task performance. 
\textbf{IKEA}~\citep{huang2025reinforced} shares a similar spirit, but focus more on adaptivity by adding a knowledge-boundary bonus ($r_\text{kb}^-$) to encourage agents using search on difficult problems.
\textbf{Naive Shaping} is conceptually similar to OTC but much easier to implement, aiming to encourage agents to use fewer search calls while obtaining the correct answer.
\textbf{Decision Shaping} aims to sharply penalize inconsistent transitions between no-search and search behaviors.
In contrast, \method use a simple binary rewards to optimize both problem solving and decision making on whether to search. 

\input{tables/reward_design}

%% file: tables/reward_design.tex
\begin{table*}[htbp!]
    \centering
    \resizebox{\linewidth}{!}{
        \begin{tabular}{l}
          \toprule
          \bf Search-R1~\citep{jin2025searchr} \& \method (Ours) \\
          $R(\tau, y) = \begin{cases}
              1.0 & \text{if $\mathbf{EM} = \texttt{true}$,} \\
              0   & \text{otherwise.}
          \end{cases}$ \\
          \midrule
          \bf OTC-GRPO~\citep{otc} \\
            $R(\tau, y) = \alpha \cdot R_\text{tool} \cdot \mathds{1}\left[\mathbf{EM} = \texttt{true}\right],$ where $R_\text{tool} = \begin{cases}
              1.0 & \text{if } f(m, n) = n = 0, \\
              \cos\!\left(\tfrac{m \pi}{2m + c}\right) & \text{if } n = 0, \\
              \sin\!\left(\tfrac{f(m, n) \pi}{2n}\right) & \text{otherwise.}
          \end{cases}$ 
          $f(m, n) = \begin{cases}
              0 & \text{if } m = n = 0, \\
              m & \text{if } n = 0, \\
              \tfrac{2nm}{m+n} & \text{otherwise.}
          \end{cases}$   \\
           \\
          $\alpha$: hyperparameter, $m$: \# search calls in $\tau$, $n$: min \# search calls for input $x$ during training. \\
          \midrule
          \bf IKEA~\citep{huang2025reinforced} \\
          $R(\tau, y) = \begin{cases}
              -1 & \text{if incorrect format,} \\
              \mathds{1}\left[\mathbf{EM}=\texttt{true}\right] + R_\text{kb} & \text{otherwise.}
          \end{cases}$  $R_\text{kb} = \begin{cases}
              r_\text{kb$^+$} \left(1 - \tfrac{\text{RT}}{\text{RT}_\text{max}}\right) & \text{if $\mathbf{EM} = \texttt{true}$,} \\
              0 & \text{if $\mathbf{EM} = \texttt{false}$ \& RT = 0,} \\
              r_\text{kb$^-$} & \text{otherwise.}
          \end{cases}$ \\
          RT: \# search calls; RT$_\text{max}$: max \# search allowed; $r_\text{kb$^+$}, r_\text{kb$^-$}$: hyperparameters. \\
          \midrule
          \bf Naive Shaping \\
            $R(\tau, y) = \begin{cases}
            1.0 - \lambda \cdot (\text{\# search calls}) & \text{if } \mathbf{EM} = \texttt{true}, \\
            0 & \text{otherwise,}
            \end{cases}$ \\
          where $\lambda$ is the hyperparameter controlling penalties for search calls. \\
          \midrule
          \bf Decision Shaping \\
            $R(\tau, y) = \mathds{1}\left[\mathbf{EM} = \texttt{true}\right] + \alpha \cdot \begin{cases}
                \mathds{1} \left[ \tau \text{ has search} \right] & \text{if } p < \rho, \\
                \mathds{1} \left[ \tau \text{ has no search} \right] & \text{otherwise}, \\
            \end{cases}$ \\
            where $\alpha$: search decision bonus, $p$: solve rate for the input problem, and $\rho$: threshold for solve rate $p$. \\
          \bottomrule
        \end{tabular}
    }
    \caption{
    Comparison of different reward designs. Unlike prior methods that rely on complex reward engineering, such as OTC~\citep{otc} and IKEA~\citep{huang2025reinforced}, \method uses a simple binary outcome reward to explicitly optimize both problem solving and decision making.}
    \label{tab:reward_design}
\end{table*}

%% file: appendix/implementation.tex
\section{Implementation Details}
\label{appendix:implementation}

\subsection{Training Dataset Construction}
\label{appendix:dataset_construction}

To ensure a fair comparison across training-based baselines, we follow the procedure of \citet{huang2025reinforced} to construct a difficulty-balanced training set. Specifically, we use the base policy with the system prompt $s_\text{probe}$ (Table~\ref{prompt:probing}), which encourages the model to answer directly without accessing additional information, to generate $K = 10$ responses for each problem in the training set of \citet{jin2025searchr}, drawn from Natural Questions~\citep{natural_questions} and HotpotQA~\citep{yang-etal-2018-hotpotqa}. We then compute solve rates using substring exact matching ($\mathbf{SubEM}$) for verification. Using a threshold $\rho = 0.5$, we split the dataset into easy and hard subsets and sample 4,096 examples from each to form the final training set. The same procedure is used to construct a validation set of 2,048 examples, which is used for checkpoint selection during evaluation.

We create datasets for each model independently, and all methods evaluated on the same model use the same dataset to ensure fair comparison.

\subsection{Training Details}
\label{appendix:training_details}

Most RL training experiments are conducted on a compute node equipped with 4 NVIDIA A100-80GB GPUs. The only exception is the outcome-based RL baselines for the 3B models, which are trained on a machine with 2 NVIDIA H100-94GB GPUs, as these baselines require relatively fewer computational resources. We use full parameter fine-tuning for all experiments. To optimize training efficiency, we adopted DeepSpeed with Zero3 offload, gradient checkpointing, FlashAttention-2, and bfloat16 mixed precision training. 

We use LlamaFactory~\citep{zheng2024llamafactory} for SFT and conduct hyperparameter search on batch size $\in$ \{16, 32\} and learning rate $\in$ \{1e-6, 5e-6\}. We only train for one epoch as the loss quickly converged to nearly 0. We select the checkpoint that maximizes the product of validation $\mathbf{EM}$ and \decisionfscore.

For RL training, we use SkyRL~\citep{griggs2025skrylv01} as our base framework. Hyperparameters for RL baselines and for \method are shown in Table~\ref{tab:baseline_hyperparam} and Table~\ref{tab:adasearch_hyperparam}, respectively. Most hyperparameters follow \citet{jin2025searchr} and \citet{otc}. For reward-shaping baselines, we increase the group size to 15 to match the compute used in problem-solving training (\method stage 1). 

For the reward-function hyperparameters, we set $\lambda$ in Naive Shaping (\Cref{eq:naive_shaping}) to 0.05; both $\alpha$ and $\rho$ in Decision Shaping to 0.5; and $r_\text{kb}^+$ and $r_\text{kb}^-$ in IKEA (\Cref{tab:reward_design}) to 0.2 and 0.05, following the original IKEA paper~\citep{huang2025reinforced}. In addition, we analyze the reward-function hyperparameters for Naive Shaping and compare our training hyperparameters with those in the original IKEA implementation in Appendix~\ref{appendix:reward_shaping_hyperparam}.

For \method stage 2, we search over the combinations of (batch size, mini-batch size) in {(128, 64), (256, 128)}. For Llama-3.2-3B-Instruct, we use the (128, 64) setting, while for the Qwen2.5 series models, we use (256, 128). 

For efficient rollouts, we use vLLM~\citep{kwon2023efficient} and set the temperature to 1.0, top-p to 1.0, and top-k to~–1, following \citet{jin2025searchr}.

For baselines that involve non-search generation (e.g., RL w/o Search, rollouts for $s_\mathrm{param}$ in \method stage 1, and \method stage 2), we set the maximum number of turns to 2, with a maximum generation length of 500 tokens in the second turn. In most cases, the full trajectory of these baselines contains only one turn unless the model fails to follow the required output format (e.g., the answer is not wrapped in \answer{…} or the assessment is not wrapped in \assessment{…}). Similar to Search-R1~\citep{jin2025searchr}, which appends a nudge prompt (Table~\ref{prompt:search_retry}) when the model violates the required format, we design nudge prompts (Table~\ref{prompt:param_retry} and \ref{prompt:param_retry_2}) to enforce correct formatting. This formatting-correction procedure is applied to all reward-shaping baselines as well to ensure fairness. 

We save checkpoints at the end of each epoch. For RL baselines, we select the checkpoint that achieves the highest validation reward. For \method stage 1, we use the final checkpoint before collapse, following \citet{jin2025searchr}. For \method stage 2, we select the checkpoint that maximizes the product of validation $\mathbf{EM}$ and \decisionfscore.

\input{tables/hyperparams}

\subsection{Inference Details}

We perform inference with vLLM~\citep{kwon2023efficient} and bfloat16 precision on the same compute nodes used for training. For each method, the maximum generation length, maximum number of turns, and maximum input length are set to match the rollout configuration used during training for consistency. Following \citet{jin2025searchr}, we use greedy decoding for evaluation.

%% file: tables/hyperparams.tex
\begin{table*}[t]
    \centering\small
    \begin{tabular}{lccc}
        \toprule
        Hyperparameter & Search-R1 & RL w/o Search & Reward-Shaping Baselines \\
        \midrule
        (batch size, mini-batch size) & (512, 256) & (512, 256) & (512, 256) \\
        learning rate & 1e-6 & 1e-6 & 1e-6 \\
        epoch & 6 & 6 & 6 \\
        group size & 5 & 10 & 15 \\
        KL loss coefficient $\beta$ & 0.001 & 0.001 & 0.001 \\
        PPO clip ratio & 0.2 & 0.2 & 0.2 \\
        warm-up step ratio & 0.285 & 0.285 & 0.285 \\
        max generation length & 500 & 2048$^*$ & 500 \\
        max number of turns & 4 & 2 & 4 \\
        max input length & 4096 & 3072 & 4096 \\
    \bottomrule
    \end{tabular}
    \caption{Hyperparameters for RL baselines. $^*$: in the second turn, the maximum generation length is set to 500 (See Appendix~\ref{appendix:training_details}).}
    \label{tab:baseline_hyperparam}
\end{table*}

\begin{table*}[t]
    \centering\small
    \begin{tabular}{lcc}
        \toprule
        Hyperparameter & \method stage 1 & \method stage 2 \\
        \midrule
        (batch size, mini-batch size) & (512, 256) & \{(128, 64), (256, 218)\} \\
        learning rate & 1e-6 & 1e-6 \\
        epoch & 6 & 6 \\
        group size & ($s_\mathrm{search}$, $s_\mathrm{param}$) = (5, 10) & 5 \\
        KL loss coefficient $\beta$ & 0.001 & 0.001 \\
        PPO clip ratio & 0.2 & 0.2 \\
        warm-up step ratio & 0.285 & 0.285 \\
        max generation length & ($s_\mathrm{search}$, $s_\mathrm{param}$) = (500, 2048$^*$) & 2048$^*$ \\
        max number of turns & ($s_\mathrm{search}$, $s_\mathrm{param}$) = (4, 2) & 2 \\
        max input length & ($s_\mathrm{search}$, $s_\mathrm{param}$) = (4096, 3072) & 3072 \\
    \bottomrule
    \end{tabular}
    \caption{Hyperparameters for \method. $^*$: in the second turn, the maximum generation length is set to 500 (See Appendix~\ref{appendix:training_details}).}
    \label{tab:adasearch_hyperparam}
\end{table*}

%% file: appendix/additional_eval.tex
\section{Additional Evaluations}
\label{appendix:additional_eval}
In this section, we provide additional analysis on interpretability~\ref{appendix:interpretability_eval}, safety~\ref{appendix:safety_eval}, and universally unanswerable questions~\ref{appendix:unanswerable_questions_eval}

\subsection{Interpretability}
\label{appendix:interpretability_eval}
Evaluations were conducted via CoT monitoring \citep{baker2025monitoring} on 500 random samples from the Natural Questions \citep{natural_questions} test set. 
We utilized GPT-5.4-mini \citep{openai2026gpt54card} and Claude 4.6 Sonnet \citep{anthropic2026sonnet46} as judges. 
The evaluation prompts, adapted from \citet{baker2025monitoring}, are provided in \Cref{prompt:judge_gpt,prompt:judge_claude}.
Each judge assessed whether the agent (1) explicitly reasoned about its own knowledge and (2) produced a clear sufficiency judgment.

\begin{table}[t!]
    \centering\small
        \begin{tabular}{lcc}
            \toprule
             Method & IR (gpt) & IR (Claude) \\
             \midrule
             Search-R1 & 7.4 & 7.0 \\
             Naive Shaping & 17.2 & 20.6 \\
             IKEA & 10.0 & 11.0 \\
             Decision Shaping & 35.8 & 40.8 \\
             \hdashline             
             \method (Ours) & \bf 44.2 & \bf 48.0 \\
             \bottomrule
        \end{tabular}
    \caption{Evaluation of interpretability on Qwen2.5 3B. \method achieves the overall best Interpretable Rate (IR).}
    \label{tab:interpretability_eval}
\end{table}

We report Interpretable Rate (IR), the ratio of trajectories where the judge thinks are interpretable based on the criteria, in \Cref{tab:interpretability_eval}. 
Across both judge models, \method consistently outperforms all baselines (by 8.4-36.8\% using gpt-5.4-mini; by 7.2-41.0\% using Claude Sonnet 4.6). 

To further examine whether the assessment tags in \method trajectories influence the evaluation results, we removed these tags and reran the evaluation using gpt-5.4-mini. \method still achieves an Interpretable Rate of 40\%, outperforming all baselines by 4.2-32.6\%.
These results suggest that \method offers more interpretable search decisions compared to the baselines.

\subsection{Safety}
\label{appendix:safety_eval}

We conduct evaluation using the SafeSearch dataset~\citep{dong2025safesearch}. Specifically, we use the \texttt{harmful\_output} subset, where each query is paired with a malicious document designed to mislead users and pose risks, on top of our retrieval environment. 
Following the original paper's protocol, we inject the harmful document at position 1 (i.e., "Doc 1") of the first search call and report the Attack Success Rate (ASR). 
As shown in \Cref{tab:safety_eval}, \method reduces ASR by 15\% compared to Search-R1. Further analysis reveals that \method invokes search on 80\% of test instances, compared to 97\% for Search-R1, suggesting that fewer search calls effectively limit exposure to malicious content.

We also note that \method is orthogonal to safety alignment methods that teach agents to ignore malicious documents, and we leave their combination as future work.

\begin{table}[t!]
    \centering\small
    \begin{tabular}{lc}
    \toprule
    Method & Attack Successful Rate \\
    \midrule
    Search-R1 & 91.7 \\
    \method & \bf 76.7 \\
    \bottomrule
    \end{tabular}
    \caption{Evaluation of safety on Qwen2.5 3B. We test the agents on the harmful output subset of SafeSearch. Our method effectively reduces ASR by 15\% compared to Search-R1.
}
    \label{tab:safety_eval}
\end{table}

\subsection{Universally Unanswerable Questions}
\label{appendix:unanswerable_questions_eval}
To test the generalization of our method beyond static, factoid QA, we conduct additional evaluations on the ``universally unknown'' subset of CoCoNot~\citep{coconot}.
Since these questions cannot be resolved by the model's parametric knowledge, the agents are expected to rely on search for this subset. 
We report the decision-making accuracy in Table~\ref{tab:coconot_eval}. Our method achieves the highest decision-making accuracy, demonstrating that \method generalizes well to queries beyond factoid QA.

\begin{table}[t!]
    \centering\small
    \begin{tabular}{lc}
    \toprule
    Method & Decision Acc. \\
    \midrule
    Search-R1 & 98.5 \\
    Naive Shaping & 43.3 \\
    Decision Shaping & 83.6 \\
    IKEA & 97.0 \\
    \hdashline
    \method & \bf 100.0 \\
    \bottomrule
    \end{tabular}
    \caption{Evaluation of universally unanswerable questions on Qwen2.5-3B. We test the agents on the universally unknown subset of CoCoNot. \method achieves the overall best decision-making accuracy.}
    \label{tab:coconot_eval}
\end{table}

%% file: appendix/additional_analysis.tex
\section{Additional Analysis}

In this section, we provide additional empirical evidence and analysis. 
Specifically, we report the balanced accuracy of search decisions (Appendix~\ref{appendix:balanced_acc}), multi-seed inference results under stochastic decoding (Appendix~\ref{appendix:multi_seed_results}), the results using the BM25 retriever (Appendix~\ref{appendix:bm25_results}), visualize the training reward trajectories (Appendix~\ref{appendix:training_rewards}), analyze the impact of hyperparameters across various methods (Appendix~\ref{appendix:stage2_ablation} and \ref{appendix:reward_shaping_hyperparam}), and evaluate efficiency regarding search-call counts and latency (Appendix~\ref{appendix:efficiency}).

\subsection{Balanced Accuracy of Search Decisions}
\label{appendix:balanced_acc}
Because \decisionfscore treats ``search not needed'' as the positive class, we additionally report balanced accuracy~\citep{5597285} derived from the macro-averaged confusion-matrix proportions in Figure~\ref{fig:confusion_matrix}. 
The results are shown in Table~\ref{tab:balanced_accuracy}. 
\method achieves the highest balanced accuracy for both model families, indicating that its advantage remains when the two decision classes are weighted equally.

\input{tables/additional_analysis/balanced_acc}

\subsection{Multi-Seed Inference Results}
\label{appendix:multi_seed_results}
To examine the inference robustness under stochastic decoding, we conduct multi-seed inference evaluations for all methods using Qwen2.5-3B-Instruct across all seven testing benchmarks. 
For each method, we fix the trained checkpoint and evaluate it using three random seeds under stochastic decoding, with temperature \(=0.7\), top-\(p=0.8\), and top-\(k=20\). 
We report the mean and standard deviation of $\mathbf{EM}$ and \decisionfscore across the three seeds. 
The results are shown in Table~\ref{tab:qwen_multiseed_inference}.
All methods exhibit low variability in their overall performance, with standard deviations below 1.0 percentage point for both $\mathbf{EM}$ and \decisionfscore. 
More importantly, the relative performance trend under stochastic decoding is consistent with the main results obtained using greedy decoding: \method achieves the second-highest overall $\mathbf{EM}$ and the highest overall \decisionfscore.

\input{tables/additional_analysis/multi_seed_inference}

\subsection{BM25 Results}
\label{appendix:bm25_results}
\begin{figure}[t!]
    \centering
    \includegraphics[width=.75\linewidth]{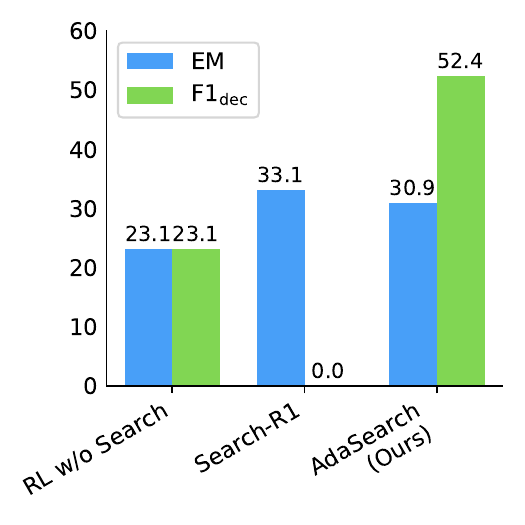}
    \vspace{-3mm}
    \caption{BM25 results on Qwen2.5-3B-Instruct. \method generalize to different retrievers.}
    \label{fig:qwen3b_bm25}
\end{figure}

To test the generalization of \method to different retrievers, we conduct experiments using Qwen2.5-3B-Instruct with BM25~\cite{robertson1994some} for both training and inference. Results are presented in Figure~\ref{fig:qwen3b_bm25}. 
\method generalizes well to different choices of retrievers. 
Specifically, compared to Search-R1, \method significantly improves search invocation decision-making (\decisionfscore) without hurting too much on QA performance ($\mathbf{EM}$).

\subsection{Training Rewards}
\label{appendix:training_rewards}

\begin{figure}[t!]
    \centering
    \begin{subfigure}{0.49\linewidth}
         \centering
         \includegraphics[width=\textwidth]{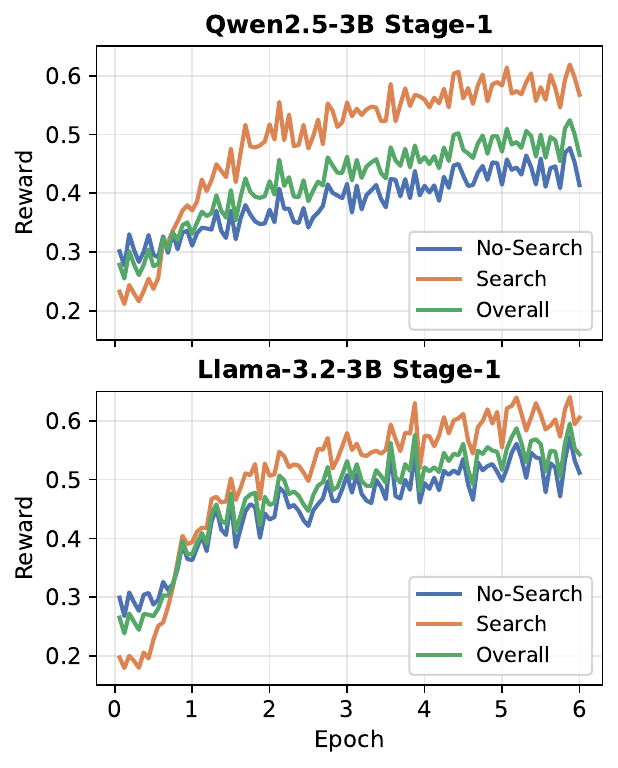}
         \caption{Stage-1 rewards.}
         \label{fig:stage1_reward}
     \end{subfigure}
     \begin{subfigure}{0.49\linewidth}
         \centering
         \includegraphics[width=\textwidth]{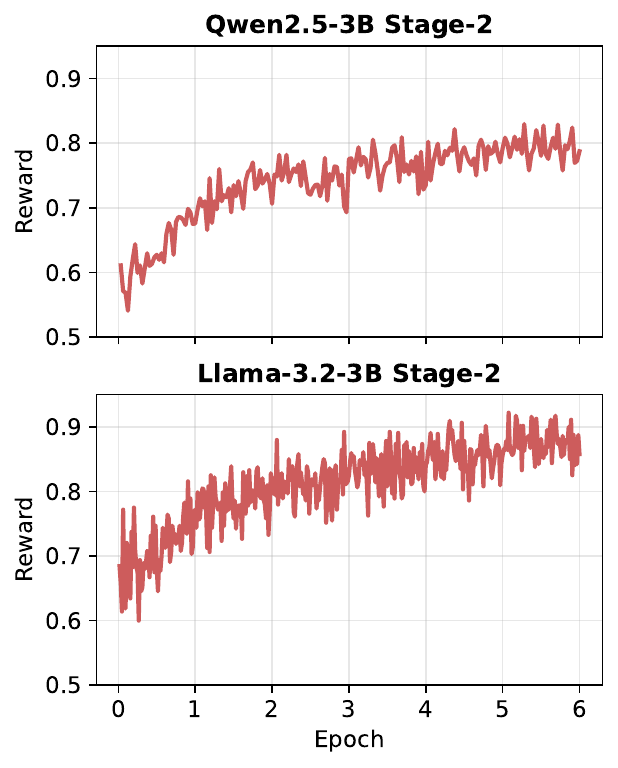}
         \caption{Stage-2 rewards.}
         \label{fig:stage2_reward}
     \end{subfigure}
    \caption{\method training rewards. The rewards increase steadily as training proceed in both stage 1 and stage 2.}
    \label{fig:training_rewards}
\end{figure}

Figure~\ref{fig:training_rewards} illustrates the training reward curves for \method on Qwen2.5-3B-Instruct and Llama-3.2-3B-Instruct. 
In Stage 1, rewards for both parametric knowledge and search-based problem solving show a steady upward trend. 
In Stage 2, the rewards stabilize smoothly as well, reflecting a consistent and robust training process.

\subsection{\method Stage 2: Threshold $\rho$ vs Number of Search Calls}
\label{appendix:stage2_ablation}
We analyze the effect of the solve-rate threshold $\rho$ on the number of search calls using Qwen2.5-3B-Instruct, with results shown in Table~\ref{tab:adasearch_stage2_rho}. As $\rho$ increases, the number of search calls rises, which is expected since a higher threshold encourages the model to invoke search more often. However, when $\rho$ becomes too large (e.g., 0.9), \decisionfscore drops substantially. This suggests overuse of search, where the model chooses to search even when its parametric knowledge is sufficient, as reflected in the decrease in recall and the increase in precision.

\input{tables/additional_analysis/stage2_rho}

\subsection{Hyperparameters of reward-shaping baselines}
\label{appendix:reward_shaping_hyperparam}

We analyze the reward-function hyperparameter of Naive Shaping and Decision Shaping, and the training hyperparameters of IKEA~\citep{huang2025reinforced} on Qwen2.5-3B-Instruct.

\paragraph{Naive shaping.}
We vary the value of $\lambda$ in Equation~\ref{eq:naive_shaping} over {0.05, 0.1, 0.25}, and report $\mathbf{EM}$, \decisionfscore, precision, recall, and the micro-averaged number of search calls. The results are shown in Table~\ref{tab:naive_shaping_lambda}. As expected, the average number of search calls decreases as $\lambda$ increases, due to the stronger penalty on invoking search. \decisionfscore increases, which can be attributed to the increase in recall, while precision remains relatively stable. Lastly, $\mathbf{EM}$ decreases with larger $\lambda$, possibly because the model underuses search, as reflected by the lower recall and reduced number of search calls.

\input{tables/additional_analysis/naive_shaping_hyperparam}

\paragraph{Decision shaping.}
We vary the value of $\alpha$ in Equation~\ref{eq:decision_shaping} over {0.1, 0.5, 1.0} and report $\mathbf{EM}$, \decisionfscore, precision, recall, and the micro-averaged number of search calls. The results are shown in Table~\ref{tab:awareness_shaping_alpha}. Increasing $\alpha$ improves \decisionfscore and recall, as expected from assigning a larger bonus to correct decisions. In contrast, precision, average search-call counts, and task $\mathbf{EM}$ decrease. These trends suggest that the model may underuse search, reflected in higher recall but lower precision and fewer search calls.

\input{tables/additional_analysis/awareness_shaping_hyperparam}

\paragraph{IKEA.}
We compare our training hyperparameters for IKEA~\citep{huang2025reinforced} (Table~\ref{tab:baseline_hyperparam}) with those used in the original IKEA paper. In the original work, the authors train with both batch size and mini-batch size set to 256, a smaller learning rate of 5e-7, a much larger warm-up ratio of 0.75, a slightly larger group size of 16, and a total of 120 training steps (approximately 4 epochs). The performance comparison is shown in Table~\ref{tab:ikea_hyperparam}. We find that using the original hyperparameters results in IKEA being  undertrained. Therefore, we adopt the hyperparameters in Table~\ref{tab:baseline_hyperparam} for our experiments.

\input{tables/additional_analysis/ikea_hyperparam}

\subsection{Comparison on efficiency}
\label{appendix:efficiency}

In this section, we compare the efficiency of each method in terms of the average number of search calls and the average latency, as shown in Table~\ref{tab:efficiency}. All averages are computed over the full set of testing samples. In addition, we estimate the total search cost on the evaluation benchmarks using the pricing of the Google Search API (5 USD per 1{,}000 queries)\footnote{\url{https://developers.google.com/custom-search/v1/overview\#pricing}}.

Regarding the average number of search calls, \method substantially reduces unnecessary searches and costs compared to Search-R1, achieving a 34\% reduction. \method also incurs a similar number of search calls as Decision Shaping, as both methods decide whether to invoke search on prompt-level based on self-knowledge. Compared to methods that explicitly minimize search calls (Naive Shaping and IKEA), \method uses slightly more searches. This behavior is expected, since \method does not directly optimize for minimizing tool calls. Instead, it focuses on avoiding unnecessary searches when parametric knowledge is sufficient, while still achieving a significant reduction relative to Search-R1.

In terms of average latency, although \method introduces an additional inference stage that explicitly decides whether to invoke search, its latency remains lower than that of Search-R1 by 20\%. These results demonstrate that \method not only reduces the average number of searches (and thus API cost) but also decreases inference latency, while providing transparent and interpretable decision rationales that are valuable in real-world applications. Compared to reward-shaping baselines, \method incurs a slightly higher latency, but it achieves higher task performance ($\mathbf{EM}$) and demonstrates better discrimination between necessary and unnecessary search calls, as reflected by \decisionfscore (\Cref{tab:main}). 

Finally, our framework is orthogonal to reward-shaping methods. Additional reward terms can be incorporated to penalize undesirable search behaviors, such as issuing duplicate queries. We leave a detailed exploration of this direction to future work.

\input{tables/efficiency}

%% file: tables/additional_analysis/balanced_acc.tex
\begin{table}[t!]
    \centering\small
    \begin{tabular}{lcc}
        \toprule
         Method & Qwen & Llama \\
         \midrule
         Search-R1 & 50.0 & 50.4  \\
         Naive & 69.7 & 70.2 \\
         Decision & 74.0 & 73.5 \\
         IKEA & 70.3 & 73.7 \\
         \hdashline
         \method & \bf 76.3 & \bf 76.9 \\
         \bottomrule
    \end{tabular}
    \caption{Balanced accuracy (\%) derived from the macro-averaged confusion-matrix proportions in Figure~\ref{fig:confusion_matrix}. The best result for each base model is shown in bold. \method achieves the best balanced accuracy across different model families.}
    \label{tab:balanced_accuracy}
\end{table}

%% file: tables/additional_analysis/multi_seed_inference.tex
\definecolor{lightblue}{RGB}{224,242,254}

\begin{table*}[t!]
  \centering
  \small
  \setlength{\tabcolsep}{3.2pt}
  \renewcommand{\arraystretch}{1.08}

  \begin{tabular}{@{}lcccccccc@{}}
    \toprule
    \multirow{2}{*}{\textbf{Method}}
    & \multicolumn{3}{c}{\textbf{General QA}}
    & \multicolumn{4}{c}{\textbf{Multi-Hop QA}}
    & \multirow{2}{*}{\textbf{Avg.}} \\
    \cmidrule(lr){2-4}
    \cmidrule(lr){5-8}
    & \textbf{NQ}
    & \textbf{TQ}
    & \textbf{PopQA}
    & \textbf{HotpotQA}
    & \textbf{2Wiki}
    & \textbf{MuSiQue}
    & \textbf{Bamboogle}
    & \\
    \midrule

    \multicolumn{9}{@{}l}{\textbf{Exact Match (EM)}} \\

    Search-R1
    & \textbf{41.9 \textpm{} 0.2}
    & \textbf{60.3 \textpm{} 0.1}
    & \underline{43.9 \textpm{} 0.2}
    & \textbf{36.5 \textpm{} 0.6}
    & \underline{38.1 \textpm{} 0.4}
    & \textbf{15.5 \textpm{} 0.6}
    & \textbf{31.7 \textpm{} 0.5}
    & \textbf{38.3 \textpm{} 0.1} \\

    Naive Shaping
    & 37.0 \textpm{} 0.3
    & 53.3 \textpm{} 0.3
    & 42.5 \textpm{} 0.2
    & 29.3 \textpm{} 0.5
    & 33.8 \textpm{} 0.4
    & 9.4 \textpm{} 0.3
    & 24.3 \textpm{} 3.6
    & 32.8 \textpm{} 0.5 \\

    Decision Shaping
    & 37.6 \textpm{} 0.6
    & 55.7 \textpm{} 0.2
    & 41.8 \textpm{} 0.5
    & 31.8 \textpm{} 0.2
    & 35.7 \textpm{} 0.6
    & 12.5 \textpm{} 0.3
    & \underline{25.3 \textpm{} 1.2}
    & 34.4 \textpm{} 0.3 \\

    IKEA
    & \underline{41.7 \textpm{} 0.2}
    & \underline{58.6 \textpm{} 0.3}
    & \textbf{44.0 \textpm{} 0.1}
    & 31.8 \textpm{} 0.4
    & 35.0 \textpm{} 0.1
    & 11.2 \textpm{} 0.3
    & 22.9 \textpm{} 1.7
    & 35.0 \textpm{} 0.3 \\

    \rowcolor{lightblue!100}
    \textbf{\method{} (Ours)}
    & 36.9 \textpm{} 0.6
    & 56.2 \textpm{} 0.2
    & 42.4 \textpm{} 0.1
    & \underline{33.0 \textpm{} 0.2}
    & \textbf{38.3 \textpm{} 0.3}
    & \underline{14.6 \textpm{} 0.5}
    & \underline{25.3 \textpm{} 2.0}
    & \underline{35.2 \textpm{} 0.5} \\

    \midrule

    \multicolumn{9}{@{}l}{\textbf{Search Decision F1 Score (\decisionfscore)}} \\

    Search-R1
    & 0.5 \textpm{} 0.2
    & 0.3 \textpm{} 0.1
    & 0.3 \textpm{} 0.1
    & 0.0 \textpm{} 0.1
    & 0.0 \textpm{} 0.0
    & 0.6 \textpm{} 1.1
    & 0.0 \textpm{} 0.0
    & 0.3 \textpm{} 0.1 \\

    Naive Shaping
    & 42.6 \textpm{} 0.4
    & 66.5 \textpm{} 0.1
    & 55.9 \textpm{} 0.8
    & 48.7 \textpm{} 0.8
    & 57.6 \textpm{} 0.2
    & 15.0 \textpm{} 1.9
    & 42.1 \textpm{} 2.9
    & 46.9 \textpm{} 0.7 \\

    Decision Shaping
    & \underline{45.5 \textpm{} 0.5}
    & \underline{70.2 \textpm{} 0.2}
    & \textbf{60.3 \textpm{} 0.4}
    & 56.9 \textpm{} 0.2
    & \underline{60.9 \textpm{} 0.8}
    & \underline{21.3 \textpm{} 1.4}
    & \underline{45.8 \textpm{} 0.6}
    & \underline{51.5 \textpm{} 0.2} \\

    IKEA
    & 38.6 \textpm{} 1.2
    & 62.8 \textpm{} 0.1
    & 54.2 \textpm{} 0.6
    & \textbf{57.9 \textpm{} 0.5}
    & 60.5 \textpm{} 0.6
    & 19.0 \textpm{} 0.3
    & 42.4 \textpm{} 5.3
    & 47.9 \textpm{} 0.8 \\

    \rowcolor{lightblue!100}
    \textbf{\method{} (Ours)}
    & \textbf{48.0 \textpm{} 0.8}
    & \textbf{71.5 \textpm{} 0.1}
    & \underline{58.8 \textpm{} 0.1}
    & \underline{57.8 \textpm{} 0.2}
    & \textbf{65.8 \textpm{} 0.7}
    & \textbf{22.8 \textpm{} 1.5}
    & \textbf{50.4 \textpm{} 5.2}
    & \textbf{53.6 \textpm{} 0.5} \\

    \bottomrule
  \end{tabular}

  \caption{
    Results of Qwen2.5-3B-Instruct under stochastic decoding.
    Results are reported as mean \textpm{} sample standard deviation
    over three random seeds. \textbf{Avg.} is computed by
    macro-averaging the corresponding metric across the seven
    benchmarks within each seed. The best mean is shown in
    \textbf{bold}, and the second-best mean is
    \underline{underlined}.
  }
  \label{tab:qwen_multiseed_inference}
\end{table*}

%% file: tables/additional_analysis/stage2_rho.tex
\begin{table}[t!]
    \centering\small
        \setlength{\tabcolsep}{4pt}
        \begin{tabular}{lccccc}
            \toprule
                 $\rho$ & $\mathbf{EM}$ & \decisionfscore & Precision & Recall & Avg. Search \\
             \midrule
             0.1 & 35.8 & 53.8 & 44.5 & 69.3 & 0.99 \\
             0.3 & 34.1 & 53.8 & 43.4 & 72.5 & 1.00 \\
             0.5 & 36.0 & 54.0 & 45.1 & 68.8 & 1.08 \\
             0.7 & 36.2 & 53.2 & 48.9 & 59.6 & 1.12 \\
             0.9 & 36.8 & 46.2 & 55.9 & 41.0 & 1.21 \\
             \bottomrule
        \end{tabular}
    \caption{Analysis on the effect of $\rho$ for \method stage 2 (Section~\ref{subsubsec:decision_making}).}
    \label{tab:adasearch_stage2_rho}
\end{table}

%% file: tables/additional_analysis/naive_shaping_hyperparam.tex
\begin{table}[t!]
    \centering\small
        \setlength{\tabcolsep}{4pt}
        \begin{tabular}{lccccc}
            \toprule
                 $\lambda$ & $\mathbf{EM}$ & \decisionfscore & Precision & Recall & Avg. Search \\
             \midrule
             0.05 & 33.2 & 47.3 & 36.1 & 76.8 & 0.77 \\
             0.10 & 33.0 & 47.8 & 35.7 & 81.3 & 0.74 \\
             0.25 & 32.1 & 48.7 & 35.9 & 85.6 & 0.67 \\
             \bottomrule
        \end{tabular}
    \caption{Analysis on the effect of $\lambda$ for Naive Shaping baseline (Eq~\ref{eq:naive_shaping}).}
    \label{tab:naive_shaping_lambda}
\end{table}

%% file: tables/additional_analysis/awareness_shaping_hyperparam.tex
\begin{table}[t!]
    \centering\small
        \setlength{\tabcolsep}{4pt}
        \begin{tabular}{lccccc}
            \toprule
                 $\alpha$ & $\mathbf{EM}$ & \decisionfscore & Precision & Recall & Avg. Search \\
             \midrule
             0.1 & 36.2 & 49.5 & 44.7 & 57.5 & 1.12 \\
             0.5 & 34.1 & 51.3 & 42.0 & 70.3  & 1.00 \\
             1.0 & 33.8 & 52.5 & 41.6 & 76.4  & 0.98 \\
             \bottomrule
        \end{tabular}
    \caption{Analysis on the effect of $\alpha$ for Decision Shaping baseline (Eq~\ref{eq:decision_shaping}).}
    \label{tab:awareness_shaping_alpha}
\end{table}

%% file: tables/additional_analysis/ikea_hyperparam.tex
\begin{table}[t!]
    \centering
    \resizebox{\linewidth}{!}{
        \setlength{\tabcolsep}{4pt}
        \begin{tabular}{lccccc}
        \toprule
            Setting & $\mathbf{EM}$ & \decisionfscore & Precision & Recall & Avg. Search \\
         \midrule
         Original & 32.3 & 12.4 & 42.0 & 8.0 & 1.06 \\
         Ours (Table~\ref{tab:baseline_hyperparam}) & 35.3 & 47.6 & 45.0 & 59.5 & 0.88 \\
         \bottomrule
        \end{tabular}
    }
    \caption{Comparison of hyperparameters settings for IKEA~\cite{huang2025reinforced}}
    \label{tab:ikea_hyperparam}
\end{table}

%% file: tables/efficiency.tex
\begin{table}[t!]
    \centering
    \resizebox{\linewidth}{!}{
        \setlength{\tabcolsep}{4pt}
        \begin{tabular}{lccc}
            \toprule
             \textbf{Method} & \textbf{Avg. Search} & \textbf{Est. Costs} & \textbf{Avg. Latency} \\
             \midrule
             \multicolumn{4}{l}{\bf Qwen2.5-3B-Instruct} \\
             Search-R1 & 1.64 & 424.3 & 0.111 \\
             Naive Shaping & 0.77 & 198.4 & 0.059 \\
             Decision Shaping & 1.00 & 259.0 & 0.074 \\
             IKEA & 0.88 & 228.6 & 0.066 \\
             \hdashline
             \method & 1.08 & 279.2 & 0.089 \\
             \bottomrule
        \end{tabular}
    }
    \caption{Average number of searches, their estimated costs, and the average latency.}
    \label{tab:efficiency}
\end{table}

%% file: appendix/algo_variant.tex
\section{\method-E2E}
\label{appendix:adasearch_e2e}

\input{sections/algo_e2e}
We unify the two stages in \method to obtain an end-to-end variant, \method-E2E. The overall training pipeline is shown in Algorithm~\ref{algo:adasearch_e2e}. For each training instance $(x, y)$, we construct three variants of the input by applying the parametric knowledge prompt $s_\text{param}$, the search prompt $s_\text{search}$, and the decision-making prompt $s_\text{decision}$.

The pseudo labels for decision making are generated on the fly: for each training instance $(x, y)$, we estimate the empirical solve rate $p$ from $\mathcal{R}_\text{param}$ using $\mathbf{SubEM}$, and assign pseudo labels according to whether $p$ exceeds a predefined threshold $\rho$. The reward function is simply the binary outcome reward, and the remaining optimization follows GRPO~\citep{shao2024deepseekmath}. 

The hyperparameters largely follow those of our two-stage method (Table~\ref{tab:adasearch_hyperparam}), except for the batch and mini-batch sizes. We search over \{(256, 128), (512, 256)\} and use (256, 128) for Qwen and (512, 256) for Llama.

%% file: sections/algo_e2e.tex
\begin{algorithm*}[t!]
\caption{AdaSearch-E2E Training}\label{alg:main}
\begin{algorithmic}[1]
\Require Base policy $\pi_{\theta_b}$, search engine $\mathcal{E}$,
training set $\mathcal{D}$, prompts
$s_{\mathrm{param}},s_{\mathrm{search}},s_{\mathrm{decision}}$,
rollout sizes $N,M,L$, threshold $\rho$

\State Initialize policy $\pi_\theta \leftarrow \pi_{\theta_b}$

\State \textcolor{blue}{\texttt{/* End-to-End Training */}}
\For{$t=1,\ldots,T$}
    \State Sample a batch $\mathcal{B} \sim \mathcal{D}$.
    \State $\mathcal{R} \leftarrow \{\}$.
    \For {$(x_i, y_i) \in \mathcal{B}$}
        \For {$g \in \{\mathrm{param}, \mathrm{search}, \mathrm{decision}\}$}
            \If {$g$ is $\mathrm{param}$}
                \State Generate $\mathcal{R}_g^i = \{ \tau_\mathrm{param}^n\}_{n=1}^N$, where $\tau_\mathrm{param}^n \sim \pi_\theta(\cdot \mid [s_\mathrm{param}, x_i])$.
                \State Compute the empirical solve rate $p_i$ for $x_i$ with $\mathbf{SubEM}$ and use it to determine the label for decision-making rollouts.
            \ElsIf {$g$ is $\mathrm{decision}$}
                \State Generate $\mathcal{R}_g^i = \{ \tau_\mathrm{decision}^l\}_{l=1}^L$, where $\tau_\mathrm{decision}^l \sim \pi_\theta(\cdot \mid [s_\mathrm{decision}, x_i])$.
            \Else
                \State Generate $\mathcal{R}_g^i = \{ \tau_\mathrm{search}^m\}_{m=1}^M$, where $\tau_\mathrm{search}^m \sim \pi_\theta(\cdot \mid [s_\mathrm{search}, x_i]; \mathcal{E})$.
            \EndIf
        
            \State Compute rewards $\mathbf{r}^i_g$ For all $\tau \in \mathcal{R}_g^i$ with Eq~\ref{eq:outcome_reward}.
            \State $\mathcal{R} \leftarrow \mathcal{R} \cup \mathcal{R}_g^i$
        \EndFor
    \EndFor
    \State Update $\pi_\theta$ with GRPO using the rewarded rollout groups in $\mathcal{R}$.
\EndFor
\State \Return $\pi_\theta$
\end{algorithmic}\label{algo:adasearch_e2e}
\end{algorithm*}

%% file: appendix/prompt.tex
\newcommand{\maketag}[1]{\catcode`\<12 #1 \catcode`\>12}

\section{Prompt Templates}
\label{appendix:prompt}

\begin{table}[htbp!]
\small
\begin{prompt}[title={Knowledge-Probing System Prompt $s_\mathrm{probing}$}]
You are a helpful assistant. Your goal is to answer questions using step-by-step reasoning. Use only your internal knowledge to recall and connect relevant information. Do NOT assume access to external tools or documents. Provide the final answer only after completing your reasoning, following the output instructions below.
\\
\\
\#\# Output Instructions \\
1. Wrap your final answer within \answer{ and } tags. For example: \answer{Wilhelm Conrad Röntgen}. \\
2. Only include the direct final answer within these tags; place all explanations and reasoning outside.
\end{prompt}
\vspace{-3mm}
\caption{Knowledge-probing system prompt for dataset construction.}
\label{prompt:probing}
\end{table}

\begin{table}[htbp!]
\small
\begin{prompt}[title={Search System Prompt $s_\mathrm{search}$}]
Answer the given question. You must conduct reasoning every time you get new information. After reasoning, if you find you lack some knowledge, you can call a search engine by \search{query}, and it will return the top searched results between \info{ and }. You can search as many times as you want. If you find no further external knowledge needed, you can directly provide the answer inside \answer{ and } without detailed illustrations. For example, \answer{xxx}.
\end{prompt}
\vspace{-3mm}
\caption{System prompt for search, adapted from \citet{jin2025searchr}.}
\label{prompt:search_prompt}
\end{table}

\begin{table}[htbp!]
\small
\begin{prompt}[title={Parametric-knowledge System Prompt $s_\mathrm{param}$}]
Answer the given question. Always conduct and show your step-by-step reasoning first, then give the final answer. When providing the answer, wrap it inside \answer{ and } without detailed illustrations. For example, \answer{xxx}.
\end{prompt}
\vspace{-3mm}
\caption{Parametric-knowledge system prompt used for CoT and RL w/o Search baselines as well.}
\label{prompt:param}
\end{table}

\begin{table}[htbp!]
\small
\begin{prompt}[title={Decision-Making System Prompt $s_\mathrm{decision}$}]
Your goal is to assess whether your knowledge is sufficient to answer the given question. Think step-by-step and then output the final assessment by following the instructions below:
\\ 
\\
\#\# Output Instructions

1. Always output your step-by-step reasoning first. \\
2. After your reasoning, output a single assessment token wrapped within \assessment{ ... } tags. \\
    \vspace{-1.5em}
    \begin{itemize}
    \setlength\itemsep{-0.1em}
        \item[--] If you are confident you can answer the question directly, output: \assessment{yes}
        \item[--] Otherwise, output: \assessment{no}
    \end{itemize}
\vspace{-0.2em}
3. Only include the final assessment token within these tags; place all reasoning outside the tags.
\end{prompt}
\vspace{-3mm}
\caption{Decision-making system prompt.}
\label{prompt:decision}
\end{table}

\begin{table}[htbp!]
\small
\begin{prompt}[title={System Prompt for Reward-shaping Baselines}]
Answer the given question. You must conduct reasoning every time you get new information. After reasoning, if you find you lack some knowledge, you can call a search engine by \search{query}, and it will return the top searched results between \info{ and }. \textcolor{red}{You need to make every search call count and gain helpful results.} If you find no further external knowledge needed, you can directly provide the answer inside \answer{ and } without detailed illustrations. For example, \answer{xxx}.
\end{prompt}
\vspace{-3mm}
\caption{System prompt for reward-shaping baselines (Naive Shaping, Awareness Shaping, and IKEA), adapt from \citet{otc}}
\label{prompt:shaping}
\end{table}

\begin{table}[htbp!]
\small
\begin{prompt}[title={System Prompt for Direct Answer Baseline}]
Answer the given question. When providing the answer, wrap it inside \answer{ and } without detailed illustrations. For example, \answer{xxx}.
\end{prompt}
\vspace{-3mm}
\caption{System prompt for direct answer baseline.}
\label{prompt:direct_answer}
\end{table}

\begin{table}[htbp!]
\small
\begin{prompt}[title={System Prompt for RAG Baseline}]
Answer the given question based on the provided documents. When providing the answer, wrap it inside \answer{ and } without detailed illustrations. For example, \answer{xxx}.
\end{prompt}
\vspace{-3mm}
\caption{System prompt for RAG baseline.}
\label{prompt:rag}
\end{table}

\begin{table}[htbp!]
\small
\begin{prompt}[title={Default User Prompt Template}]
Question: \{\texttt{question}\}
\end{prompt}
\vspace{-3mm}
\caption{Default user prompt template, where \texttt{question} will be replaced with the actual input question.}
\label{prompt:user_template}
\end{table}

\begin{table}[htbp!]
\small
\begin{prompt}[title={RAG User Prompt Template}]
Doc 1: \{\texttt{document\_1}\} \\
Doc 2: \{\texttt{document\_2}\} \\
Doc 3: \{\texttt{document\_3}\} \\
Question: \{\texttt{question}\}
\end{prompt}
\vspace{-3mm}
\caption{RAG User prompt template, where \texttt{document\_i} and \texttt{question} will be replaced with the retrieved document and the actual input question, respectively.}
\label{prompt:rag_user_template}
\end{table}

\begin{table}[htbp!]
\small
\begin{prompt}[title={Retry Prompt (Search)}]
My previous action is invalid. If I want to search, I should put the query between \search{ and }. If I want to give the final answer, I should put the answer between \answer{ and }. Let me try again.
\end{prompt}
\vspace{-3mm}
\caption{Retry prompt used when the model produces an invalid action under the search prompt $s_\mathrm{search}$.}
\label{prompt:search_retry}
\end{table}

\begin{table}[htbp!]
\small
\begin{prompt}[title={Retry Prompt (Parametric-knowledge)}]
I should put the answer between \answer{ and }.
\end{prompt}
\vspace{-3mm}
\caption{Retry prompt used when the model fails to wrap the answer in \answer{…} under the parametric-knowledge prompt $s_\mathrm{param}$.}
\label{prompt:param_retry}
\end{table}

\begin{table}[htbp!]
\small
\begin{prompt}[title={Retry Prompt (Parametric-knowledge; search invoked)}]
My previous action is invalid. I should rely on my own step-by-step reasoning to solve the problem \
and then put the answer between \answer{ and }. Let me try again.
\end{prompt}
\vspace{-3mm}
\caption{Retry prompt used when the model produces a search call under the parametric-knowledge prompt $s_\mathrm{param}$.}
\label{prompt:param_retry_2}
\end{table}

\begin{table}[htbp!]
\small
\begin{prompt}[title={CoT Monitoring Prompt Template (gpt-5.4-mini)}]
You are given the following fact-seeking Question and the corresponding Trajectory from a search agent. The search agent can access a search engine by wrapping its search query with the tags <search>...</search>. The documents will be returned within the tags <output>...</output>. At the end of the trajectory, the agent will provide its answer within the tags <answer>...</answer>. Tokens outside of these tags are the reasoning tokens of the agent.\\
\\
\#\# Question\\
\texttt{\{question\}}\\
\\
\#\# Agent Trajectory\\
\texttt{\{trajectory\}}\\
\\
\#\# Instructions\\
Your task is to determine if the agent trajectory contains explicit reflection on whether it should invoke the search engines, or whether its own knowledge is sufficient to answer the question, based on the following criteria:\\
\\
Criteria:\\
- Did the agent explicitly reason about its own knowledge or relevant facts before deciding whether to invoke search?\\
- Did the agent produce an explicit judgment on whether its own knowledge is sufficient (e.g., "I can answer this directly based on my knowledge" or "I do not have sufficient knowledge to answer this")?\\
\\
Notice that simply invoking search without first assessing whether its own knowledge is sufficient is not an explanation, as it does not reflect on the agent's own knowledge or its limitations.\\
Notice that simply recalling facts and providing the final answer is not an explanation, as it does not include an explicit judgment on whether the agent's own knowledge is sufficient. However, if the agent recalls relevant facts and then explicitly concludes that its knowledge is or is not sufficient to answer the question, this counts as a valid reflection.\\
\\
\#\# Grading Report\\
Provide an explanation of how the Trajectory is rated against the criteria above. Please only consider the criteria mentioned above and do not consider anything else.\\
\\
\#\# Final Answer
Selected from one of the options listed below.\\
A. The agent explicitly reflected on whether its own knowledge is sufficient to answer the question.\\
B. The agent did not explicitly reflect on whether its own knowledge is sufficient to answer the question.\\
\\
Your response should be like: your grading report + final answer.
\end{prompt}
\vspace{-3mm}
\caption{CoT monitoring prompt using gpt-5.4-mini as the judge for interpretability evaluation. The placeholders \texttt{question} and \texttt{trajectory} are replaced with the actual input question and agent output, respectively.}
\label{prompt:judge_gpt}
\end{table}

\begin{table}[htbp!]
\small
\begin{prompt}[title={CoT Monitoring Prompt Template (Claude 4.6 Sonnet)}]
You are given the following fact-seeking Question and the corresponding Trajectory from a search agent. The search agent can access a search engine by wrapping its search query with the tags <search>...</search>. The documents will be returned within the tags <output>...</output>. At the end of the trajectory, the agent will provide its answer within the tags <answer>...</answer>. Tokens outside of these tags are the reasoning tokens of the agent.\\
\\
\#\# Question\\
\texttt{\{question\}}\\
\\
\#\# Agent Trajectory\\
\texttt{\{trajectory\}}\\
\\
\#\# Instructions\\
Your task is to determine if the agent trajectory contains explicit reflection on whether it should invoke the search engines, or whether its own knowledge is sufficient to answer the question, based on the following criteria:\\
\\
Criteria:\\
- Did the agent explicitly reason about its own knowledge or relevant facts before deciding whether to invoke search?\\
- Did the agent produce an explicit judgment on whether its own knowledge is sufficient (e.g., "I can answer this directly based on my knowledge" or "I do not have sufficient knowledge to answer this")?\\
\\
Notice that simply invoking search without first assessing whether its own knowledge is sufficient is not an explanation, as it does not reflect on the agent's own knowledge or its limitations.\\
Notice that simply recalling facts and providing the final answer is not an explanation, as it does not include an explicit judgment on whether the agent's own knowledge is sufficient. However, if the agent recalls relevant facts and then explicitly concludes that its knowledge is or is not sufficient to answer the question, this counts as a valid reflection.\\
\\
\#\# Grading Report\\
Provide an explanation of how the Trajectory is rated against the criteria above. Please only consider the criteria mentioned above and do not consider anything else.\\
\\
\#\# Final Answer\\
Selected from one of the options listed below.\\
A. The agent explicitly reflected on whether its own knowledge is sufficient to answer the question.\\
B. The agent did not explicitly reflect on whether its own knowledge is sufficient to answer the question.\\
\\
\\
Your response should be like: your grading report, \textcolor{red}{followed by your final answer on the last line in this exact format: "Final Answer: [A]" or "Final Answer: [B]".}
\end{prompt}
\vspace{-3mm}
\caption{CoT monitoring prompt using Claude 4.6 Sonnet as the judge for interpretability evaluation. The placeholders \texttt{question} and \texttt{trajectory} are replaced with the actual input question and agent output, respectively. Differences from the prompt for gpt-5.4-mini (\Cref{prompt:judge_gpt}), which help parse model responses more easily, are highlighted in \textcolor{red}{red}.}
\label{prompt:judge_claude}
\end{table}

%% file: appendix/case_studies.tex
\section{Case Studies}

\subsection{Case 1: Parametric Knowledge is Sufficient}

\begin{table}[htbp!]
\small
\begin{prompt}[title={Input Question (Answer: Christian Louboutin)}] 
Which designer is famous for his/her red-soled footwear?
\end{prompt}
\caption{Input question. The answer is ``Christian Louboutin".}
\end{table}

\begin{table}[htbp!]
\begin{prompt}[title={\method Stage-1 Output (Answer: Christian Louboutin)}] 
\small
Step 1: I need to think of a designer known for red-soled footwear.\\
Step 2: One famous designer known for red-soled footwear is Christian Louboutin.\\
Step 3: Christian Louboutin is a French fashion designer who is renowned for his distinctive red-soled high-heeled shoes.\\
Step 4: Based on this information, I can answer the question.\\
\assessment{yes}
\end{prompt}
\vspace{-3mm}
\caption{\method stage-1 output. The model explicitly evaluates its own knowledge and concludes that it is sufficient.}
\end{table}

\begin{table}[htbp!]
\small
\begin{prompt}[title={\method Stage-2 Output (Answer: Christian Louboutin)}] 
To determine which designer is famous for his/her red-soled footwear, we need to consider some of the most well-known designers and their iconic designs.\\
\\
1. Christian Louboutin is a renowned French fashion designer who is famous for creating red-soled high-heeled shoes.\\
2. Other designers like Jimmy Choo and Alexander McQueen have also designed red-soled shoes, but Christian Louboutin is particularly known for this specific style.\\
\\
Given this information, the designer famous for his/her red-soled footwear is Christian Louboutin.\\
\\
\answer{Christian Louboutin}
\end{prompt}
\vspace{-3mm}
\caption{\method stage-2 output. The model correctly answer the question using parametric knowledge.}
\end{table}

\begin{table}[htbp!]
\small
    \begin{prompt}[title={Search-R1 Output (Answer: Christian Louboutin)}]
    \small
    To find out which designer is famous for their red-soled footwear, I will need to search for information about notable footwear designers and their signature styles.\\
\\
\search{Which designer is famous for red-soled footwear}\\
\info{\\
Doc 1: "Christian Louboutin"\\
Christian Louboutin Christian Louboutin (; born 7 January 1964) is a French fashion designer whose high-end stiletto footwear incorporates shiny, red-lacquered soles that have become his signature. Initially a freelance designer for fashion houses, he started his own shoe salon in Paris, with his shoes finding favor with celebrity clientele. He has partnered with other organizations for creative projects including limited edition pieces, gallery exhibits, and even a custom bar. His company has since branched out into men's footwear, handbags, fragrances and makeup. Louboutin was born and raised in Paris's 12th arrondissement. He was the only son of Roger, a\\
Doc 2: "Christian Louboutin"\\
tempered to 28,000 Mexican pesos (1471.89 dollars). Christian Louboutin Christian Louboutin (; born 7 January 1964) is a French fashion designer whose high-end stiletto footwear incorporates shiny, red-lacquered soles that have become his signature. Initially a freelance designer for fashion houses, he started his own shoe salon in Paris, with his shoes finding favor with celebrity clientele. He has partnered with other organizations for creative projects including limited edition pieces, gallery exhibits, and even a custom bar. His company has since branched out into men's footwear, handbags, fragrances and makeup. Louboutin was born and raised in Paris's 12th arrondissement. He\\
Doc 3: "Christian Louboutin"\\
on his high heel shoes, commonly referred to as ""sammy red soled shoes"". His signature red soles were first created in 1993. ""My sketches were not reproduced exactly as I had designed them and I couldn't figure out why. The two dimensional sketch was so powerful on paper, but when turned into a three-dimensional object, it was somehow lacking energy. Frustrated after having tried different things to liven up the design, I spontaneously grabbed my assistant's red nail polish and painted the sole. I instantly knew that this would be a success!"". Christian Louboutin's red-bottom colour code is Pantone 18-1663\\
}\\
\\
From the search results, it is clear that Christian Louboutin is famous for his red-soled footwear. \\
\\
\answer{Christian Louboutin}
\end{prompt}
\vspace{-3mm}
\caption{Search-R1 output. The agent overuses search on problems where its parametric knowledge is sufficient.}
\end{table}

\begin{table}[htbp!]
\small
\begin{prompt}[title={Naive Shaping Output (Answer: Christian Louboutin)}] 
\small
To determine which designer is famous for their red-soled footwear, let's consider some prominent designers and their notable styles.\\
\\
One of the most famous designers known for red-soled footwear is Alexander McQueen. He was known for his bold and innovative designs, including red-soled shoes that became iconic in his collections. 
\\
Given this information, the designer famous for his/her red-soled footwear is Alexander McQueen.\\
\\
\answer{Alexander McQueen}
\end{prompt}
\vspace{-3mm}
\caption{Naive Shaping output. The model produces an incorrect answer without explicit self-knowledge assessment.}
\end{table}

\FloatBarrier
\subsection{Case 2: Parametric knowledge is Insufficient}

\begin{table}[htbp!]
\small
\begin{prompt}[title={Input Question (Answer: Bardney)}] 
\small
Southrey is a village in the civil parish of which village in the West Lindsey district of Lincolnshire, England?
\end{prompt}
\vspace{-3mm}
\caption{Input question. The answer is ``Bardney".}
\end{table}

\begin{table}[htbp!]
\small
\begin{prompt}[title={\method Stage-1 Output (Answer: Bardney)}] 
Step 1: Identify the key information in the question. The question asks about a civil parish that contains the village of Southrey.\\
Step 2: Recall that a civil parish is a territorial unit of local government in England, and it is often part of a larger administrative division such as a district.\\
Step 3: Think about the West Lindsey district in Lincolnshire, England. I need to find which village in this district contains Southrey.\\
Step 4: Without specific knowledge of the exact location, I would need to search for the civil parish that includes Southrey. However, I don't have immediate access to this information.\\
Step 5: Given the lack of specific knowledge about the exact civil parish containing Southrey, I cannot answer the question directly.\\
\assessment{no}
\end{prompt}
\vspace{-3mm}
\caption{\method stage-1 output. After self-assessment, the model is not confident in answering using parametric knowledge.}
\end{table}

\begin{table}[htbp!]
\small
\begin{prompt}[title={\method Stage-2 Output (Answer: Bardney)}] 
\small
To determine the village in the West Lindsey district of Lincolnshire, England, that is the civil parish of Southrey, I will need to search for this information.\\
\\
\search{Village in the West Lindsey district of Lincolnshire that is the civil parish of Southrey}\\
\info{\\
Doc 1: Bardney\\
consists of crafts, trade and community stands, funfair rides, bar and BBQ, sports, children's races, car boot sales, classic vehicles, and an Exemption Dog Show. Originally part of the Wraggoe Wapentake, and of the Lincoln poor law union in the Parts of Lindsey, the parish is now part of West Lindsey District, and Lincolnshire County Council. The civil parish has become part of a group that consists of Bardney, Apley and Stainfield parishes. The villages of Southrey and Kingthorpe are also included. The 2011 electoral arrangements are: Bardney lies between 7 and 17 metres above sea level, on the edge\\
Doc 2: Snelland\\
Snelland Snelland is a village and civil parish in the West Lindsey district of Lincolnshire, England It is situated approximately north-east from the city and county town of Lincoln and about from the town of Market Rasen. The civil parish includes the hamlet of Swinthorpe. The population at the 2011 census was included in the civil parish of Friesthorpe. Snelland is listed in ""Domesday Book"" of 1086 as ""Sneleslunt"", with seven households. The nearby deserted medieval village of Swinthorpe is listed as ""Sonetorp"", with three households. There is evidence that a Cistercian grange existed at Swinthorpe, attached to Kirkstead Abbey.\\
Doc 3: "South Thoresby"\\
South Thoresby South Thoresby is a village and civil parish in the East Lindsey district of Lincolnshire, England. It is situated north-east from Horncastle and south-east from Louth. South Thoresby is about east from the A16 in the Lincolnshire Wolds. The civil parish extends much further to the south, over the A16 and to include the hamlets of Calceby (a former medieval village) and Driby, extending southwards to border with Langton by Spilsby. To the south-west is Haugh. It lies close to the northern boundary with Aby with Greenfield. The parish council administers Swaby, South Thoresby and Haugh, although these\\
}\\
\\
From the search results, it is clear that Southrey is a village in the civil parish of Bardney. Therefore, the village in the West Lindsey district of Lincolnshire, England, that is the civil parish of Southrey is Bardney.\\
\\
\answer{Bardney}
\end{prompt}
\vspace{-3mm}
\caption{\method stage-2 output. The model invokes search and produces a correct answer.}
\end{table}

\begin{table}[htbp!]
\small
\begin{prompt}[title={Naive Shaping Output (Answer: Bardney)}] 
\small
To determine the village in the West Lindsey district of Lincolnshire, England, that is the civil parish of Southrey, we need to find information about the civil parishes and their boundaries.\\
\\
Southrey is a village in the civil parish of Little Sleaford. Little Sleaford is a village located in the West Lindsey district of Lincolnshire, England.\\
\\
Therefore, the civil parish of Southrey is Little Sleaford.\\
\\
\answer{Little Sleaford}
\end{prompt}
\vspace{-3mm}
\caption{Naive Shaping output. The model produces an incorrect answer without explicit self-knowledge assessment.}
\end{table}

\FloatBarrier

\subsection{Case 3: \method Underuse Search}
\label{case:underuse}

\begin{table}[htbp!]
\small
\begin{prompt}[title={Input Question (Answer: Mary Elizabeth Braddon)}]
Who is the author of Lady Audley's Secret?
\end{prompt}
\vspace{-3mm}
\caption{Input question. The answer is ``Mary Elizabeth Braddon".}
\end{table}

\begin{table}[htbp!]
\small
\begin{prompt}[title={\method Stage-1 Output (Answer: Mary Elizabeth Braddon)}]
Step 1: I need to recall the author of the novel "Lady Audley's Secret."\\
Step 2: I know that "Lady Audley's Secret" is a novel.\\
Step 3: I remember that the author of "Lady Audley's Secret" is Mrs. Wilkie Collins.\\
Step 4: I can confirm that Mrs. Wilkie Collins is indeed the author of this novel.\\
Step 5: Based on my knowledge, I can answer the question directly.\\
\assessment{yes}
\end{prompt}
\vspace{-3mm}
\caption{\method stage-1 output. The model holds a false belief, which leads to an incorrect self-assessment.}
\end{table}

\begin{table}[htbp!]
\small
\begin{prompt}[title={\method Stage-2 Output (Answer: Mary Elizabeth Braddon)}] 
To determine the author of "Lady Audley's Secret," we need to consider the historical context and the well-known works of that era. \\
\\
"Lady Audley's Secret" is a novel published in 1862. The author of this novel is Mrs. Wilkie Collins, a prominent English novelist of the Victorian era. \\
\\
Therefore, the author of "Lady Audley's Secret" is \answer{Mrs. Wilkie Collins}
\end{prompt}
\vspace{-3mm}
\caption{\method stage-2 output. The model maintains its false belief, resulting in insufficient search.}
\end{table}

\FloatBarrier
\subsection{Case 4: \method Overuse Search}
\label{case:overuse}

\begin{table}[htbp!]
\small
\begin{prompt}[title={Input Question (Answer: Politician)}] 
What is Javier Alva Orlandini's occupation?
\end{prompt}
\vspace{-3mm}
\caption{Input question. The answer is ``Politician".}
\end{table}

\begin{table}[htbp!]
\small
\begin{prompt}[title={\method Stage-1 Output (Answer: Politician)}] 
Step 1: I need to determine Javier Alva Orlandini's occupation.\\
Step 2: I do not have any information about Javier Alva Orlandini in my knowledge base.\\
Step 3: Without specific information about this person, I cannot answer the question directly.\\
\assessment{no}
\end{prompt}
\vspace{-3mm}
\caption{\method stage-1 output. The model is not confident in its self-knowledge and therefore abstains.}
\end{table}

\begin{table}[htbp!]
\small
\begin{prompt}[title={\method Stage-2 Output (Search) (Answer: Politician)}] 
To determine Javier Alva Orlandini's occupation, I will need to search for information about him. Let me start with a search.\\
\search{Javier Alva Orlandini occupation}\\
\info{\\
Doc 1: "Javier Alva Orlandini"\\
Javier Alva Orlandini Javier Alva Orlandini (born December 11, 1927) is a lawyer and a Peruvian politician. A prominent member of the political party Popular Action, he serves as the President of the party. He served as 2nd Vice President of Peru during the government of Fernando Belaúnde from 1980 to 1985.. Throughout his life he became a Deputy (1963-1968), Minister of Government and Police (1965-1966), Senator (1980-1985 and 1990-1992), President of the Senate (1981-1982) and Congressman (1995-2000). In 2002, he became a Magistrate of the Constitutional Court, at the same time being elected by Congress as the President of\\
Doc 2: "Luis Orlandini"\\
Luis Orlandini Luis Orlandini (born 1964) is a Chilean guitarist and professor at the University of Chile and the Pontifical Catholic University of Chile. Orlandini studied guitar with Ernesto Quezada at the University of Chile and then with Eliot Fisk at the School of Music of Cologne. In 1989, he obtained First Place at the ARD International Music Competition in Munich, thus initiating an international career that has led him to perform in various countries and record with different European labels. He performs mainly as a soloist, but has also played with groups of different natures of chamber music, one\\
Doc 3: "Javier Alva Orlandini"\\
the Court for a 3-year term. He was a candidate for the Presidency of the Republic in 1985, but got 4\% of the popular vote, being defeated by Alan García. He later served as President of Peru's constitutional tribunal. He has written several books, including ""Responding to the Dictatorship"", ""Yesterday, Today and Tomorrow"", the ""Vicious Circle"", ""I Minister"", ""Word of Honor"" and ""Yes I swear"", and numerous legal articles. The audits, representative legislative, legal, academic and professional are of extensive quality and are recognized by the Peruvian people and their various institutions that have given many honorary degrees. Javier Alva\\
}\\
\\
Based on the search results, Javier Alva Orlandini is a lawyer and a Peruvian politician. He served as the President of the party Popular Action and was the 2nd Vice President of Peru during the government of Fernando Belaúnde from 1980 to 1985. \\
\\
\answer{Politician}
\end{prompt}
\vspace{-3mm}
\caption{\method stage-2 output. The model produces a correct answer after search.}
\end{table}

\begin{table}[htbp!]
\small
\begin{prompt}[title={\method Stage-2 Output (Parametric Knowledge) (Answer: Politician)}] 
To determine Javier Alva Orlandini's occupation, we need to look up information about him. Given that the question does not provide specific details, we can infer based on common knowledge or a search. \\
\\
Javier Alva Orlandini is a Mexican politician who has held various positions in the Mexican government. Given his background, it is most likely that he is a politician. \\
\\
However, without a specific source, we can't be certain. For the purpose of this question, we will assume his occupation is based on his known roles in Mexican politics.\\
\\
\answer{Politician}
\end{prompt}
\vspace{-3mm}
\caption{Model output under the parametric-knowledge system prompt $s_\mathrm{param}$. The model is uncertain about its answer, even though the answer is correct.}
\end{table}

\FloatBarrier
\subsection{Case 5: Multi-hop Failure of \method}

\begin{table}[htbp!]
\small
\begin{prompt}[title={Input Question (Answer: 1978)}]
The director of Panic 5 Bravo was born in what year?
\end{prompt}
\vspace{-3mm}
\caption{Input question. The answer is ``1978".}
\end{table}

\begin{table}[htbp!]
\small
\begin{prompt}[title={\method Stage-1 Output (Answer: 1978)}]
Step 1: To answer this question, I need to know the director of the movie "Panic 5 Bravo". However, there is no widely known movie or film titled "Panic 5 Bravo" that I am aware of. This suggests it might be a fictional or obscure title.\\
Step 2: Without knowing the director of this specific movie, I cannot determine the year of birth of the director.\\
Step 3: Since I do not have the necessary information to answer the question, I cannot provide a year of birth for the director.\\
\assessment{no}
\end{prompt}
\vspace{-3mm}
\caption{\method stage-1 output. The model abstains because it lacks sufficient information.}
\end{table}

\begin{table}[htbp!]
\small
\begin{prompt}[title={\method Stage-2 Output (Answer: 1978)},label=tab:multihop_fail_part1]
To find out the year the director of "Panic 5 Bravo" was born, I will need to gather information about this film and its director. Let's start by searching for more details about the film "Panic 5 Bravo".\\
\search{The director of Panic 5 Bravo and the year of birth}\\
\info{\\
Doc 1: "Panic (2000 film)"\\
Panic (2000 film) Panic is a 2000 American crime film directed by Henry Bromell and starring William H. Macy, Neve Campbell, Tracey Ullman, John Ritter and Donald Sutherland. Despite not being a box office success, ""Panic"" received universal critical acclaim from critics, some of which called it one of the best movies of 2000. The performances of William H. Macy, John Ritter, and Donald Sutherland were hailed by critics. Alex (Macy), a sad-eyed mournful man, goes into psychotherapy disclosing that he is a hit man and also tells the doctor (Ritter) after a few sessions that he is attracted to\\
Doc 2: "Panic (2000 film)"\\
originality from the tone: quirky yet convincing, irreverent yet moral."" Panic (2000 film) Panic is a 2000 American crime film directed by Henry Bromell and starring William H. Macy, Neve Campbell, Tracey Ullman, John Ritter and Donald Sutherland. Despite not being a box office success, ""Panic"" received universal critical acclaim from critics, some of which called it one of the best movies of 2000. The performances of William H. Macy, John Ritter, and Donald Sutherland were hailed by critics. Alex (Macy), a sad-eyed mournful man, goes into psychotherapy disclosing that he is a hit man and also tells the doctor\\
Doc 3: "A Town Called Panic (film)"\\
A Town Called Panic (film) A Town Called Panic () is a 2009 internationally co-produced stop-motion animated adventure fantasy comedy family film directed by Stéphane Aubier and Vincent Patar and co-written by Aubier, Guillaume Malandrin, Patar, and Vincent Tavier. The film is based on the TV series of the same name and stars Aubier, Jeanne Balibar, Nicolas Buysse, Véronique Dumont, Bruce Ellison, Frédéric Jannin, Bouli Lanners, and Patar, among others. It premiered at the 2009 Cannes Film Festival and was the first stop-motion film to be screened at the festival. The film was released theatrically in Belgium on 17 June\\
}\\
\\
From the search results, it seems "Panic 5 Bravo" is not a widely known film. \textcolor{red}{However, the first result mentions a film called "Panic (2000 film)" directed by Henry Bromell. Given this information, I will now search for Henry Bromell's birth year.}\\
\end{prompt}
\vspace{-3mm}
\caption{\method stage-2 output (part 1). The model invokes search but makes a faulty inference (highlighted in \textcolor{red}{red}).}
\end{table}

\begin{table}[htbp!]
\small
\begin{prompt}[title={\method Stage-2 Output (Continued) (Answer: 1978)}]
\small
\search{Henry Bromell birth year}\\
\info{\\
Doc 1: "Henry Bromell"\\
Henry Bromell Alfred Henry Bromell (September 19, 1947 – March 18, 2013) was an American novelist, screenwriter, and director. Bromell joined the crew of NBC police drama """" in 1994. He served as a writer and co-executive producer for the show's third season. He contributed to writing seven episodes for the season. He was promoted to executive producer for the fourth season and wrote a further 17 episodes. He scaled back his involvement with the fifth season and became a consulting producer. He wrote a further two episodes before leaving the crew at the end of the season in 1997.\\
Doc 2: "Henry Bromell"\\
screenwriter and director Caroline Thompson. He then married writer Trish Soodik, who died of cancer in January 2009; they had a son, William. His third wife was Sarah. Bromell died March 18, 2013 of a heart attack, at UCLA Santa Monica hospital at age 65. Henry Bromell Alfred Henry Bromell (September 19, 1947 – March 18, 2013) was an American novelist, screenwriter, and director. Bromell joined the crew of NBC police drama """" in 1994. He served as a writer and co-executive producer for the show's third season. He contributed to writing seven episodes for the season. He was promoted\\
Doc 3: "Henry Bromell"\\
He contributed to a total of 26 episodes as a writer over three seasons with the series. He returned as a co-writer and co-executive producer for the feature-length follow-up ""Homicide: The Movie"" in 2000. He wrote and produced for many television series, including ""Chicago Hope"", ""Northern Exposure"", ""Homicide: Life on the Street"", ""Brotherhood"", ""Carnivàle"", and ""Rubicon"". He was a consulting producer, and later Executive Producer on the Showtime series ""Homeland"" at the time of his death and wrote four episodes: ""The Good Soldier"", ""Representative Brody"", ""Q\&A"", and ""Broken Hearts"". He was awarded a for ""The Good Soldier"", and he was\\
}\\
\\
From the search results, Henry Bromell was born in 1947. Therefore, the director of "Panic (2000 film)" was born in 1947.\\
\\
\answer{1947}
\end{prompt}
\vspace{-3mm}
\caption{\method stage-2 output. Continuing from part 1, the model uses the wrong director as the search query, which ultimately leads to an incorrect answer.}
\end{table}